\documentclass[journal]{IEEEtran}
\usepackage[numbers,sort&compress,square]{natbib}

\usepackage{times}
\usepackage{epsfig}
\usepackage{graphicx}
\usepackage{amsmath}
\usepackage{amssymb}

\usepackage{subfigure}
\usepackage{stackrel}
\usepackage{wasysym}
\usepackage{dsfont}
\usepackage{bm}
\usepackage{booktabs}
\usepackage{multirow}
\usepackage{epstopdf}
\usepackage{color,xcolor,array,hyperref}

\makeatletter

\newcommand*{\@rowstyle}{}

\newcommand*{\rowstyle}[1]{
  \gdef\@rowstyle{#1}%
  \@rowstyle\ignorespaces%
}

\newcolumntype{=}{
  >{\gdef\@rowstyle{}}%
}

\newcolumntype{+}{
  >{\@rowstyle}%
}

\makeatother

\begin{document}
%
\title{A Probabilistic Quality Representation Approach to Deep Blind Image Quality Prediction}
%
%
%

\author{Hui Zeng,
        Lei Zhang,~\IEEEmembership{Senior Member,~IEEE}
        Alan C. Bovik,~\IEEEmembership{Fellow,~IEEE}
\thanks{\IEEEcompsocthanksitem H.~Zeng and L.~Zhang are with the Department of Computing, The Hong Kong Polytechnic University. \protect
(E-mail: cshzeng@comp.polyu.edu.hk, cslzhang@comp.polyu.edu.hk).}
\thanks{\IEEEcompsocthanksitem Alan C. Bovik is with the Department of Electrical and Computer Engineering,
The University of Texas at Austin, Austin, TX 78712 USA (E-mail: bovik@ece.utexas.edu).}.
\thanks{}}

\maketitle

\begin{abstract}
Blind image quality assessment (BIQA) remains a very challenging problem due to the unavailability of a reference image.
Deep learning based BIQA methods have been attracting increasing attention in recent years, yet it remains a difficult task to train a robust deep BIQA model because of the very limited number of training samples with human subjective scores. Most existing methods learn a regression network to minimize the prediction error of a scalar image quality score. However, such a scheme ignores the fact that an image will receive divergent subjective scores from different subjects, which cannot be adequately represented by a single scalar number. This is particularly true on complex, real-world distorted images. Moreover, images may broadly differ in their distributions of assigned subjective scores. Recognizing this, we propose a new representation of perceptual image quality, called probabilistic quality representation (PQR), to describe the image subjective score distribution, whereby a more robust loss function can be employed to train a deep BIQA model. The proposed PQR method is shown to not only speed up the convergence of deep model training, but to also greatly improve the achievable level of quality prediction accuracy relative to scalar quality score regression methods. The source code is available at \url{https://github.com/HuiZeng/BIQA_Toolbox}.

\end{abstract}

\begin{IEEEkeywords}
Blind image quality assessment, convolutional neural network, deep learning, image quality representation.
\end{IEEEkeywords}

\section{Introduction}
\label{sec:intro}

With the explosion of visual media data, huge amounts of digital images are generated, stored, processed and transmitted every day. During these different stages, the images can undergo diverse, often multiple distortions, arising from under-/over-exposure, various blurs, noise corruption, compression artifacts, and so on.  Developing algorithms that can automatically monitor the perceptual quality of images is not only crucial to improve user experience, but also important for the design of image processing algorithms and devices, such as digital cameras \cite{bovik2013automatic}.
In many practical applications, it is very hard, if not impossible, to obtain a reference image of the image to be assessed, making powerful full-reference \cite{wang2004image,wang2003multiscale,sheikh2006image,zhang2011fsim} and reduced-reference \cite{wang2011reduced} image quality assessment (IQA) methods inapplicable. Thus it has become increasingly important to develop effective no-reference, or more practically, blind IQA (BIQA) methods which can predict image quality without any additional information.

\begin{figure}[tb]
\centering
\subfigure{
\begin{minipage}[b]{0.9\linewidth}
\centering
\includegraphics[width=1.0\textwidth]{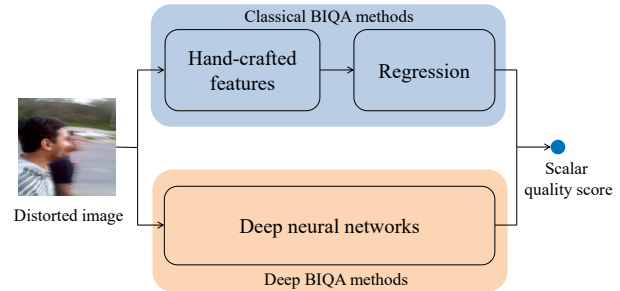}
\end{minipage}}%
\caption{Flowchart of existing BIQA methods. Both the classical and the deep BIQA methods directly train a model to regress the scalar quality scores.}
\label{figure:flowchart}
\end{figure}

BIQA methods aim to predict the scalar quality score of the image in a manner that is consistent with human opinions. Most existing BIQA methods follow the flowchart shown in Fig. \ref{figure:flowchart}. Classical BIQA methods \cite{moorthy2010two,moorthy2011blind,ye2012unsupervised,mittal2012no,tang2011learning,xue2014blind,liu2014no,ghadiyaram2016perceptual,xu2016blind} typically first extract some handcrafted features (e.g., derived from natural scene statistics models) to represent the distorted image, and then train a regression model (e.g., by support vector regression (SVR)) to map the feature representations to subjective quality scores. An obvious limitation of those BIQA methods is that the handcrafted features may not be powerful enough to adequately represent complex image structures and distortions, and therefore, to predict perceptual image quality accurately enough.
Recently, deep learning techniques have achieved great successes in solving various image recognition and processing problems \cite{krizhevsky2012imagenet}. The remarkable capability of deep neural networks to learn discriminative features provides a very promising option for addressing the challenging BIQA task.
Several attempts \cite{kang2014convolutional,tang2014blind,hou2015blind,kim2017fully,chen2016recurrent} have been made to apply deep learning technology to the BIQA tasks, yet clear successes have been elusive.
The main reason for this is that the success of deep learning relies heavily on large-scale annotated data, like the image recognition oriented ImageNet dataset \cite{imagenet_cvpr09}; unfortunately, for the task of BIQA, there do not yet exist any databases containing sufficient quantities of training images which have associated human subjective quality scores.

Since the subjective scoring of images quality is very expensive and cumbersome, existing IQA databases can only provide very limited numbers of images on which subjective quality scores have been collected. Indeed, the most popular and representative legacy IQA databases, such as LIVE IQA \cite{LIVE_database}, CSIQ \cite{larson2010most} and TID2013 \cite{ponomarenko2013color}, generally contain no more than 3,000 distorted images, usually generated from no more than 30 source images, hence they can only include very limited degrees of content variations, implying a poor representation of the high dimensional image space. In this direction, the authors of \cite{ghadiyaram2016massive} created a database of about 1200 authentically distorted natural images that is much larger in content
 variations and distortion types. We will refer to this LIVE In-the-Wild Image Quality Challenge Database as ``LIVE Challenge."
By conducting an online  crowdsourced picture quality study, they obtained more than 350,000 human subjective scores on these images. Even this number of training samples is inadequate to train a robust deep model having millions of parameters. However, given the unprecedented number of picture contents and the very wide diversity of distortions and combinations of distortions, LIVE Challlenge presents a difficult test even for pre-trained very deep learned IQA models \cite{SPM_Kim}.

Researchers have also used a variety of data augmentation methods to generate more training samples \cite{SPM_Kim}. The most popular data augmentation method is to extract a huge number of small image patches to train patch-based models. This technology has proven very effective on several synthetically distorted image quality databases which simulate clearly defined homogeneous, single impairments \cite{kang2014convolutional}. Unfortunately, the perceptual qualities of local image patches generally differ from each other and from the perceived quality of the entire image, especially for complex authentic distortions which are often not homogeneous. Other data augmentation methods, such as generating proxy quality scores using full-reference IQA methods to replace or supplement human opinions \cite{xue2013learning,ma2017waterloo,kim2017fully,ma2017dipiq}, or using ranked image pairs \cite{ma2017dipiq}, cannot be applied on practical BIQA applications where reference image information is not available.

Towards making progress on this problem, we have attempted to develop better ways to train a robust deep BIQA model using only a limited number of annotated images.
Our approach to this problem exploits the fact that the perceptual qualities of real-world images are highly subjective, and a given authentically distorted image may be assigned very different quality scores by different human subjects. For example, the average standard deviation of the subjective scores of the images in the LIVE Challenge database is 19.27 on a Mean Opinion Score (MOS) scale of [0,100] \cite{ghadiyaram2016massive}. Heretofore, this property has not been discussed or utilized, likely since the simpler, legacy databases of singular, synthetic distortions contain much tighter ranges of reported quality opinions, as might be expected. Hence, existing learning-based BIQA methods have only used the MOS to represent image quality when training regression models, ignoring the potentially useful and predictive information contained in the distributions of perceptual quality opinion scores.

Thus we have developed and describe here a new image quality representation scheme that captures the distributions of the often diverse subjective opinions of images. Since the original subjective opinions assigned to each image are generally unavailable, we propose a probabilistic quality representation (PQR) that we use to approximately describe the subjective score distribution of each image. As described in detail further in Sec. \ref{sec:proposed method}, we do this by first defining a set of quality ``anchors," then transform each original scalar MOS into a vectorized PQR, based on the distance between the MOS value and each quality anchor value. By using the PQR to augment the subjective information contained in the training samples, we are able to define a more robust loss function to use when training deep BIQA models. The results of our experiments show that this strategy not only speeds up the training process, but also make it possible to achieve much higher prediction accuracy than widely used scalar quality score regression methods.

\section{Related work}
\label{sec:relatedwork}

Existing BIQA methods can be conveniently divided into two categories: classical regression based models and more recent deep learning based methods.
In classical regression based BIQA methods, a set of handcrafted features usually are extracted first to capture quality-related aspects of a set of distorted training images, then, a regression model is learned that maps the image representations onto scalar quality scores. Along this line, Moorthy \textit{et al.} \cite{moorthy2010two} proposed a two-step framework called BIQI, and used it to create the first BIQA model based on a model of natural scene statistics (NSS). BIQI operates by using a support vector classifier (SVC) to first identify the likely distortion types, then a trained SVR model to predict the distortion-specific image quality. The BRISQUE model developed by Mittal \textit{et al.} \cite{mittal2012no} computes the scene statistics of locally normalized luminance coefficients expressed in the spatial domain, then uses them as input to train an SVR to make quality predictions. Tang \textit{et al.} \cite{tang2011learning} employed three sets of low level features (natural image statistics, texture measures and blur/noise measures) to train an SVR model on each group of features. Liu \textit{et al.} \cite{liu2014no} computed spatial and spectral image entropies under an NSS model to train an SVR to make quality predictions. Ye \textit{et al.} \cite{ye2012unsupervised} employed a codebook based image representation using soft-assignment and max-pooling techniques, and used it to train a linear SVR model to map the code-words onto subjective scores. Ghadiyaram \textit{et al.} \cite{ghadiyaram2016perceptual} proposed the FRIQUEE model, which exploits a bag of NSS features feeding an SVR that was trained to perform image quality prediction.

Recently, a few deep learning based methods have been developed to solve the BIQA problem, achieving promising performance on the legacy IQA databases. Similar to classical methods, these deep models operate by mapping distorted images or image patches onto scalar quality scores. Kang \textit{et al.} \cite{kang2014convolutional} trained a shallow CNN model to perform BIQA, where small image patches are fed into a 5 layer CNN using a score regression loss. Tang \textit{et al.} \cite{tang2014blind} introduced a semi-supervised rectifier neural network that conducts BIQA. They first trained two layers of restricted Boltzmann machines
(RBMs) based on a set of hand-crafted features, then fine-tuned the pre-trained model on human labels using a kernel regression function. To alleviate the overfitting problem, Kim \textit{et al.} \cite{kim2017fully} first pre-trained a model on a large number of image patches with proxy quality labels generated by a state-of-the-art full-reference method, then aggregated the feature vectors of image patches into image level representations to predict image-wise quality scores.
Although there are some differences in the training strategies used, all of the deep methods mentioned above learn to directly regress the scalar quality scores.

Since existing IQA databases contain insufficient numbers of images, data augmentation is currently necessary to train a deep BIQA model. The most popular data augmentation strategy is to extract a huge number of small image patches in order to train a patch-based quality prediction model. This technique has proven effective on legacy IQA databases containing synthetic distortions. However, the quality scores of small image patches can vary greatly across an image owing to spatial inhomogeneities of both content and of authentic distortions. Since human judgements of image patch qualities are exceedingly difficult to obtain, replacements for human scores have been considered. Several authors \cite{xue2013learning,ma2017waterloo,kim2017fully,ma2017dipiq} have employed state-of-the-art full-reference IQA algorithms to generate proxy scores on many image patches. An obvious limitation of this technique is the requirement of reference image, which is not usually available. Ma \textit{et al.} \cite{ma2017dipiq} proposed to train a deep BIQA model using a large number of ranked image pairs. However, in this case, the rank order is also difficult to know in practical applications where no reference image is available.

Here, we define a new path to the ``deep IQA problem" by defining a new way to represent perceptual image quality, which we call probabilistic quality representation (PQR). It is based on the observation that different human viewers may experience different perceptions of the quality of distorted images. While it is often believed that people generally agree with respect to
the quality of a picture, this may become less true as the distortions become more variable and authentic. This suggests that image quality could be represented as a (vectorized) probability, instead of as a scalar quantity. The idea is to embody richer image quality information in the PQR, making it possible to train a more stable deep BIQA model even while using a limited number of training samples.

\section{Proposed method}
\label{sec:proposed method}
We begin by first briefly outlining the limitations of existing deep BIQA methods, and then we motivate and describe the PQR framework. This leads to a probabilistic representation of distorted images and the deep BIQA model training process.

\subsection{Framework of the Proposed Method}

\begin{figure}[tb]
\centering
\subfigure{
\begin{minipage}[b]{1.0\linewidth}
\centering
\includegraphics[width=1.0\textwidth]{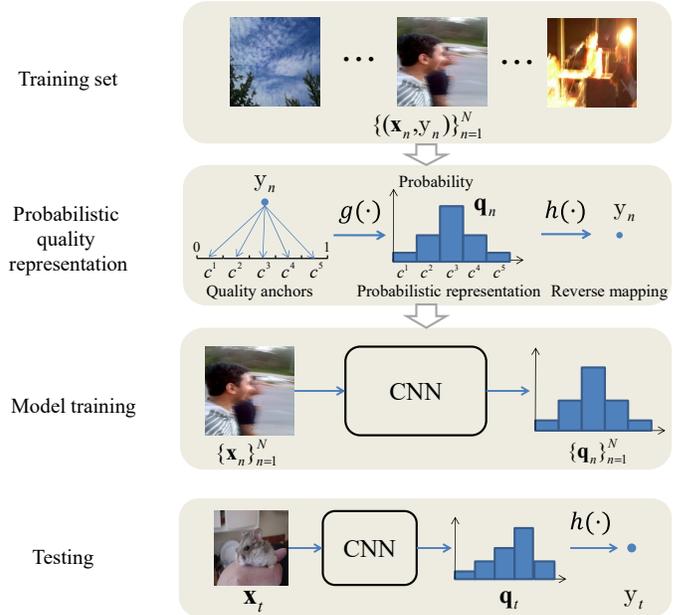}
\end{minipage}}
\caption{Framework of the proposed method. Given a training database, we first define a set of quality ``anchors" lying within the range of subjective scores, then transform the scalar quality scores into a PQR vector via a transformation function $g(\cdot)$. Simultaneously, we learn another function $h(\cdot)$ which accurately maps each vectorized PQR back to a scalar quality score.
A CNN model is then trained on the PQRs of a set of training samples. In the testing stage, the trained CNN model
outputs a PQR vector for the input image, which is then mapped to a scalar quality score using the learned
function $h(\cdot)$.}
\label{figure:overview}
\end{figure}

Without loss of generality, we will only consider the patch-wise training (with a fixed patch size such as $224 \times 224$) although our method could easily be applied on a whole-image basis. Given a database with subjective scores, patch-wise training methods extract image patches for data augmentation. Denote by $\{(\mathbf{x}_{n},y_{n})\}_{n=1}^{N}$ the training set, where $y_{n}$ is the quality score of the $n$-th training sample $\mathbf{x}_{n}$ and $N$ is the total number of training samples. In prior patch-based methods, $y_{n}$ is usually inherited from the MOS values assigned to whole images. The few existing deep BIQA methods \cite{kang2014convolutional,tang2014blind,hou2015blind,kim2017fully,SPM_Kim} mostly employ the following objective function to learn a regression model:
\begin{equation}\label{equ:regression}
\min_{\omega}\frac{1}{N}\sum_{n=1}^{N}\|f_{\omega}(\mathbf{x}_{n})-y_{n}\|^{p}
\end{equation}
where $f_{\omega}(\cdot)$ is a deep model with parameter $\omega$, which takes the training sample $\mathbf{x}_{n}$ as input, and when trained, outputs a scalar quality prediction. Usually the squared Euclidean distance (i.e. $p=2$) is employed in Eq. (\ref{equ:regression}) because of its easy optimization.

The objective function (Eq. (\ref{equ:regression})) is thus used to optimize a deep model having millions of parameters to map high-dimensional input image data $\mathbf{x}_{n}$ onto scalar quality scores $y_{n}$. However, as mentioned before, existing IQA databases are only able to provide very limited numbers of images that are supplied with subjective quality scores. As pointed out in \cite{SPM_Kim}, obtaining subjective image quality scores is a more difficult and time-consuming task than, for example, ImageNet class labels \cite{imagenet_cvpr09}. Without enough labeled training samples, it is hard to train a stable, convergent deep model \cite{geman1992neural}, and a model trained on a small dataset may not generalize well to unseen data. Extracting a huge number of small patches from a limited number of images is one way to partly alleviate the data shortage, but it requires the use of proxy patch quality scores in the absence of subjective patch labels. Using whole-image subject scores as proxy patch labels can improve training stability to some extent, but it also introduces the issue of quality bias, since the true subjective qualities of image patches are quite inhomogeneous over space, and can differ greatly from the subjective reports of whole-image quality, especially those afflicted by real-world, authentic distortions. The use of full-reference IQA predictions computed by engines like FSIM \cite{zhang2011fsim}, MS-SSIM \cite{wang2003multiscale}, or VIF \cite{sheikh2006image} as proxy patch scores is also fraught, since these models are also imperfect. Moreover, they require references patches, implying that they may only be used to a deep model on synthetic distortions. The use of proxy subjective scores generated by no-reference IQA algorithms like BRISQUE \cite{mittal2012no} or similar models \cite{moorthy2010two,moorthy2011blind,ye2012unsupervised,tang2011learning,xue2014blind,liu2014no,ghadiyaram2016perceptual,xu2016blind} is also highly questionable. While these models can be trained on authentically distorted images like those in LIVE Challenge, the use of a very shallow model to train a deep model in highly questionable. Overall, the benefits brought by patch based data augmentation have not been significant.

In order to train a more robust deep BIQA model, even when only using a limited set of training samples, we propose the use of a new probabilistic quality representation (PQR) model that can be used to enrich the subjective information associated with each training sample, thereby endowing a more informative loss function. The subjective quality of a given image cannot be adequately described by a single scalar value, because the intra-subjective opinions of any distorted image may vary widely, especially for inhomogeneous authentic distortions that may co-exist in complex combinations. Moreover, the empirical distribution of subjective opinions, whether recorded or not, may vary both within the span of an image as well as across images. Thus it would be useful to have available a informative probabilistic representation of possible perceptual quality levels.

A visual flow diagram illustrating the framework of our approach is shown in Fig. \ref{figure:overview}. Given a training database, we first define or deduce a set of quality ``anchors" that lie within the score range, then transform the scalar quality scores into a PQR vector via a transformation function $g(\cdot)$. A transformation function $h(\cdot)$ is simultaneously learned, which accurately converts a vector PQR back to a scalar quality score. Given the PQRs of a set of training samples as inputs, an end-to-end CNN model can be trained. During the testing stage, the trained CNN model outputs a PQR vector descriptive of input image, which can then be mapped to a scalar quality score using the learned function $h(\cdot)$.

\subsection{Probabilistic Quality Representation (PQR)}

We define three key components of our proposed PQR: the quality anchors, probabilistic representation mapping and the reverse mapping to scalar scores.

\textbf{Quality anchors.} The first step is to find $M$ quality anchors that fall within the overall range of quality scores. The quality anchors are intended to serve as a discrete set of model score realizations shared by all images, onto which a model probabilistic representation can be defined. There are many ways to define the anchors. The simplest is to divide the numerical range of possible subjective scores into a small number of equally spaced intervals. For example, one could partition the subjective quality range into five Likert-type levels representing ``bad,'' ``poor," ``fair," ``good," and ``excellent" \cite{LIVE_database,ghadiyaram2016massive}. These natural divisions may already be available as part of the subjective dataset being used. We divide the score range, typically [0, 1] or [0, 100],  into $M$ equal bins, then define the midpoints $\{c^{m}\}_{m=1}^{M}$ of the bins as the quality anchors. This approach is unsupervised, and does not require using any information from the training data.

Supervised methods, which utilize information from the overall per-image subjective data of a database, can also be used to determine the quality anchors. Since the per-image quality scores are scalar values, the optimal Lloyd-Max quantization scheme \cite{DBLP:journals/tit/Lloyd82,max1960quantizing} is a good candidate. A Lloyd-Max quantizer applied to a training corpus of per-image subjective quality scores will find an optimal set of $M$ quality anchors representing all of the training samples by minimizing the mean-square quantization error. Specifically, given a set of $N$ training scores $\{y_{n}\}_{n=1}^{N}$ and a fixed number of quantization levels $M$, the Lloyd-Max quantizer finds the optimal decision boundaries $\{b^{m}\}_{m=1}^{M-1}$ and quality anchors $\{c^{m}\}_{m=1}^{M}$ minimizing:
\begin{equation}\label{equ:Lloyd-Max}
\min_{b^{m},c^{m}}\sum_{m=1}^{M}\sum_{y_{n}\in[b^{m-1},b^{m})}^{}(y_{n}-c^{m})^2.
\end{equation}
While this approach is not based on patch scores (although it could be were they available), it does utilize the scores taken over a substantial (training) portion of an entire subjective quality dataset, and may be viewed as broadly representative of the subjective quality distribution of that dataset.

\textbf{Probabilistic representation.} Given a set of quality anchors, the scalar quality scores are then mapped into vectorized PQRs; that is, each image is assigned a set of probabilities of the quality anchors. The PQR is designed under the following two constraints: 1) Given an image with assigned MOS $y_{n}$, define a set of probabilities $q_{n}^{m}$ associated with the anchors $\{c^{m}\}_{m=1}^{M}$, such that $q_{n}^{m}$ is large when the Euclidean distance $\|y_{n}-c^{m}\|^{2}$ is small, and decreases monotonically with increasing distance; 2) The per-image anchor probabilities sum to 1.
A simple and effective function is the soft-mapping function $g(\cdot,m)$:
\begin{equation} \label{equ:socre transformation}
q_{n}^{m}=g(y_{n},m)=\frac{\exp(-\beta\|y_{n}-c^{m}\|^{2})}{\sum_{i=1}^{M}\exp(-\beta\|y_{n}-c^{i}\|^{2})},  m=1,2,...,M,
\end{equation}
where $q_{n}^{m}$ is the probability that the $n$-th training sample belongs to the $m$-th quality anchor, and $\beta$ is a scaling constant. In our implementations, we normalized $y_{n}$ to the range [0,1] on all examined databases, and determined a common value of $\beta$ over all databases.
The PQR mapping (Eq. (\ref{equ:socre transformation})) is motivated in a similar way as soft-assignment methods that are commonly used in clustering algorithms \cite{kearns1998information,van2010visual}. The squared Euclidean distance is convenient but not necessary; we also tested other distance metrics, including the $l_{1}$-norm distance, but found that the final quality prediction results depend little on this choice.

\textbf{Reverse mapping.} By transforming the scalar image quality score into a PQR vector, a deep BIQA model is learned which will output PQR vectors descriptive of perceptual image quality. This  probabilistic description is of great interest and may be used in a variety of ways, e.g., to train a more robust deep BIQA  model. Since the basic evaluation criteria (e.g., SRCC and PLCC) of modern BIQA models are generally computed on produced scalar quality predictions, it is desirable to be able to re-map the output PQR vectors back to scalar quality scores.  This is a vector to scalar transform problem, which we solve by learning a regression function $h(\cdot)$ that maps PQR vectors back to scalar quality scores, specifically, by minimizing the following error function:
\begin{equation}\label{equ:back transformation}
err = \frac{1}{N}\sum_{n=1}^{N}\|h(\mathbf{q}_{n}) - y_{n}\|^{2}
\end{equation}
where $\mathbf{q}_{n}$ is the vector form of $\{q_{n}^{m}\}_{m=1}^{M}$. This reverse mapping is a relatively simple regression task (with $M$ independent variables and $1$ dependent variable) that is easily and accurately solved using a linear SVR model. We have found the average absolute error $\frac{1}{N}\sum_{n=1}^{N}|h(\mathbf{q}_{n}) - y_{n}|$ of this minimization to be smaller than 0.01 on a MOS scale of [0, 1]) for reasonable choices of $\beta$ and $M$, which are the only model parameters other than those defining the deep network.

\subsection{Training the Deep BIQA Model}

\textbf{Loss function.}
Given the PQR vectors of an adequate set of training samples, we are able to train a deep BIQA model to conduct image quality prediction. As we will show, the enriched probabilistic quality descriptions contained in the PQR vectors lead to more robust quality predictions, even without a very large corpus of labeled training data. Since the transformed probabilities $\mathbf{q}_{n}$ lie in the range $[0,1]$ and sum to 1, we also employ a softmax layer to ensure that the output of the deep model satisfies the same properties.
Denote by $\mathbf{\widetilde{q}}_{n}$ the output of the softmax layer. Since both the output of the deep model $\mathbf{\widetilde{q}}_{n}$ and the target $\mathbf{q}_{n}$ are probability distributions, it is a natural and effective choice to minimize the Kullback-Leibler (KL) divergence between these two probability distributions:
\begin{equation} \label{equ:KL divergence}
D_{KL}(\mathbf{q}_{n}\|\mathbf{\widetilde{q}}_{n}) = \sum_{m=1}^{M}q_{n}^{m}\log\frac{q_{n}^{m}}{\widetilde{q}_{n}^{m}}.
\end{equation}
Since the target probability distribution $\mathbf{q}_{n}$ is fixed, minimizing the KL divergence is identical to minimizing the cross-entropy \cite[Chap. 6.9]{bishop1995neural}. Our final loss function is then:
\begin{equation}\label{equ:loss}
\min_{\omega}\frac{1}{N}\sum_{n=1}^{N}\sum_{m=1}^{M}-q_{n}^{m}\log\widetilde{q}_{n}^{m}
\end{equation}

Training deep BIQA models using our proposed PQR model exploits some attractive optimization properties that are not shared by traditional scalar quality regression. As mentioned, the PQR supplies much richer information descriptive of the subjective opinions of the training samples. This increased information richness when training and applying the deep BIQA models leads to increased accuracy. Further, the use of the softmax cross-entropy loss enforces stronger constraints on the highly flexible deep CNN models that we deploy, thereby accelerating and stabilizing the training process, while also supplying better generalization capability.

During the testing stage, the PQR prediction vector of each image patch is re-mapped to a scalar quality score using the learned function $h(\cdot)$ in Eq. (\ref{equ:back transformation}). Finally, the scalar quality scores over all of the image patches are pooled, yielding a whole-image quality prediction score. Here, we simply average the patch scores (average pooling), obtaining exceptional performance in doing so. However, there is ample scope for improving the pooling process, especially given the rich set of probabilistic spatial quality representations that are still available in the form of patch PQR predictions. One can easily envision adding additional output layers to optimize the many-to-one pooling function.

\textbf{CNN models.}
We fine tuned several well-known pre-trained deep CNN architectures on the IQA databases to evaluate the efficacy of our proposed PQR model. For a more comprehensive evaluation, we also trained a shallow CNN network using smaller image patches, and compared its performance with that of the deep CNN networks.

Specifically, we finetune two well-known deep CNN networks, AlexNet \cite{krizhevsky2012imagenet} and ResNes50 \cite{resnet}\footnote{Other popular CNN architectures, such as GoogLeNet \cite{szegedy2015going} and the VGG models \cite{simonyan2014very}, lie between these two models in terms of their depth and complexity. Hence we did not evaluate the performance of PQR on them.}. Both of these models were first pre-trained on the ImageNet \cite{imagenet_cvpr09} database, to conduct an image classification task, ostensibly learning very general image feature representations that can be transferred, with additional training, to conduct picture quality prediction. The specific configurations of the pre-trained networks can be found in the original papers. The image input sizes are constrained to $227 \times 227$ for AlexNet and $224 \times 224$ for ResNet50. To enable fine-tuning, we replaced the last fully-connected (FC) layers of each pre-trained network with a new FC layer, followed by the loss function layer. To alleviate overfitting, we add a single dropout layer with dropout rate equal to 0.5 immediately before the last FC layer.

\begin{figure}[tb]
\centering
\subfigure{
\begin{minipage}[b]{0.9\linewidth}
\centering
\includegraphics[width=1.0\textwidth]{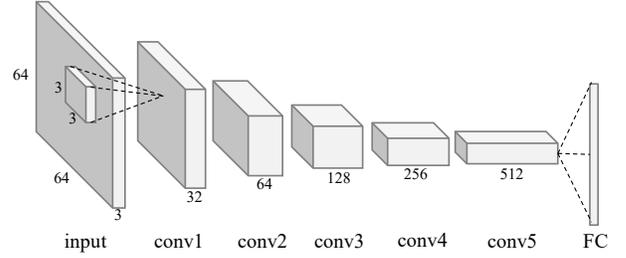}
\end{minipage}}
\caption{Configuration of the compared shallow S\_CNN model.}
\label{figure:S_CNN}
\end{figure}

Since many existing BIQA methods \cite{kang2014convolutional,kim2017fully,ma2017dipiq} train a shallow CNN model using small image patches, we also evaluated our proposed PQR method using a shallow CNN model (hereafter referred to as S\_CNN). The overall configuration of S\_CNN is shown in Fig. \ref{figure:S_CNN}. It consists of 5 convolutional (conv) layers followed by a single FC layer and takes $64\times64$ RGB image patches as input. All of the conv-layers employ filters of size $3\times3$ with stride 1, except for the last conv-layer, which uses $2\times2$ filters. Each conv-layer is followed by a max-pooling layer and a rectified linear unit (ReLU) activation layer. All of the max-pooling layers use $2\times2$ kernels with stride 2. After each downsampling, the number of filters in the conv-layer is doubled. No padding is used throughout the network in order to reduce the number of parameters. The dropout layer is placed just before the FC layer. The final S\_CNN model requires the optimization of only 0.9 million parameters, as compared with AlexNet (62 millions) and ResNet50 (26 millions).

\begin{table*}[t]
\centering
\caption{Summary of the databases employed in the experiments.}
\label{table:summary of databases}
\begin{tabular}{|c|c|c|c|c|}
\hline
Databases       & Reference image number   & Distorted image number   & distortion types   & Authentic/Synthetic    \\\hline
LIVE Challenge \cite{ghadiyaram2016massive} & N.A. & 1,162 & Numerous & Authentic \\\hline
LIVE IQA \cite{LIVE_database} & 29 & 779 & 5 & Synthetic \\\hline
CSIQ \cite{larson2010most} & 30 & 866 & 6 & Synthetic  \\\hline
TID2013 \cite{ponomarenko2013color} & 25 & 3,000 & 24 & Synthetic   \\\hline
\end{tabular}
\end{table*}

\section{Experiments}

\subsection{Databases}

We tested the PQR model on four representative IQA databases: LIVE Challenge \cite{ghadiyaram2016massive}, LIVE IQA \cite{LIVE_database}, CSIQ \cite{larson2010most} and TID2013 \cite{ponomarenko2013color}. A summary of these databases is reported in Table \ref{table:summary of databases}. The LIVE Challenge database is currently the largest IQA database along several important dimensions, including the number of distinct image contents (nearly 1,200, as compared to less than 3 dozens in the other databases) and the diversity of distortions and distortion combinations, which are essentially as numerous as the contents. It is also the only database to contain authentic, real-world distortions, since the images were captured by a wide variety of mobile camera devices, by numerous photographers, under highly diverse conditions. More than 350,000 subjective scores were collected via an online crowdsourced human study. MOS is provided for all the images in LIVE Challenge.

LIVE IQA, which was first introduced in 2003, was the first successful public-domain IQA database. It contains 29 reference images and 779 distorted ones, each impaired by one of four levels of five types of synthetic distortions: JPEG2000 (JP2K) compression, JPEG compression, additive white noise (WN), Gaussian blur (GB) and simulated fast fading channel distortion (FF). The differential Mean Opinion Scores (DMOS) of all of the distorted images are provided. The CSIQ database consists of 866 distorted images, simulated on 30 reference images. Six synthetic distortion types were used: JPEG, JP2K, WN, GB, additive pink Gaussian noise, and global contrast decrements. The DMOS is provided for all distorted images. The TID2013 database contains the largest number of distorted images (3,000), synthetically generated on 25 reference images using 24 synthetic distortion types, each at five degradation levels. The MOS of all the distorted images is provided.

\subsection{Experimental Setup}

When fine tuning the pre-trained AlexNet and ResNet50 models, we randomly extracted 50 image crops (of sizes $227 \times 227$ for AlexNet and $224 \times 224$ for ResNet50) from each training image, except on TID2013, we extracted 25 crops per image since this database contains more distorted images. All of the image crops inherited the PQR of the source image. The fine-tuning process iterated for 20 epochs, using a batch size of 256 for AlexNet, and 10 epochs with a batch size of 64 for ResNet50. The learning rate was set to be a logarithmically spaced vector in the interval [1e-3,1e-4] for both models\footnote{This setting follows the example in the MatConvNet toolbox \cite{vedaldi15matconvnet}.}.
When training S\_CNN, we extracted 500 image patches (of size $64 \times 64$) per image on each database (except 250 patches on TID2013). The network parameters of S\_CNN were initialized using the method proposed in \cite{he2015delving}, and the training process was allowed to iterate for 40 epochs using a batch size of 1024. The learning rate was set to be a logarithmically spaced vector in the interval [1e-2,1e-3].

In the testing stage, we extracted overlapped image patches at a fixed stride (64 for the fine tuned deep models and 32 for the shallow S\_CNN model) from each testing image. Denser crops were not found to bring obvious improvements in our experiments. The PQR prediction vector of each image crop was mapped to a scalar quality score using the pre-trained linear SVR model. Average pooling was used to output a final whole-image quality score.

Two metrics, Spearman's rank correlation coefficient (SRCC) and Pearson's linear correlation coefficient (PLCC), were used to evaluate the performances of the learned BIQA models. On each database, we randomly divided the samples into a training set and a testing (or validation) set without overlap in image content. All the experiments were repeated 10 times and the median SRCC and PLCC were reported as the final results. The MatConvNet toolbox \cite{vedaldi15matconvnet} was used to train the CNN models on a PC equipped with a GTX 1080Ti. The SVR model was trained using the LIBSVM toolbox \cite{CC01a}.

\subsection{Selection of Parameters $\beta$ and $M$}

There are two free parameters in the PQR model: the smoothing parameter $\beta$ and the anchor quantization level $M$ in Eq. (\ref{equ:socre transformation}). We selected the two parameters via experimental evaluation on the four databases. On each database 60\% of the images were used for training, 20\% of the images were used for validation, and the remaining 20\% of images were used for evaluation. We only used the AlexNet for the parameter selection study, since the PQR is independent of the CNN model used. The optimized values of $\beta$ and $M$ were then applied to all of the CNN models in the experiments.

\begin{figure*}[t]
\centering
\subfigure{
\begin{minipage}[t]{0.4\linewidth}
\centering
\includegraphics[width=1.0\textwidth]{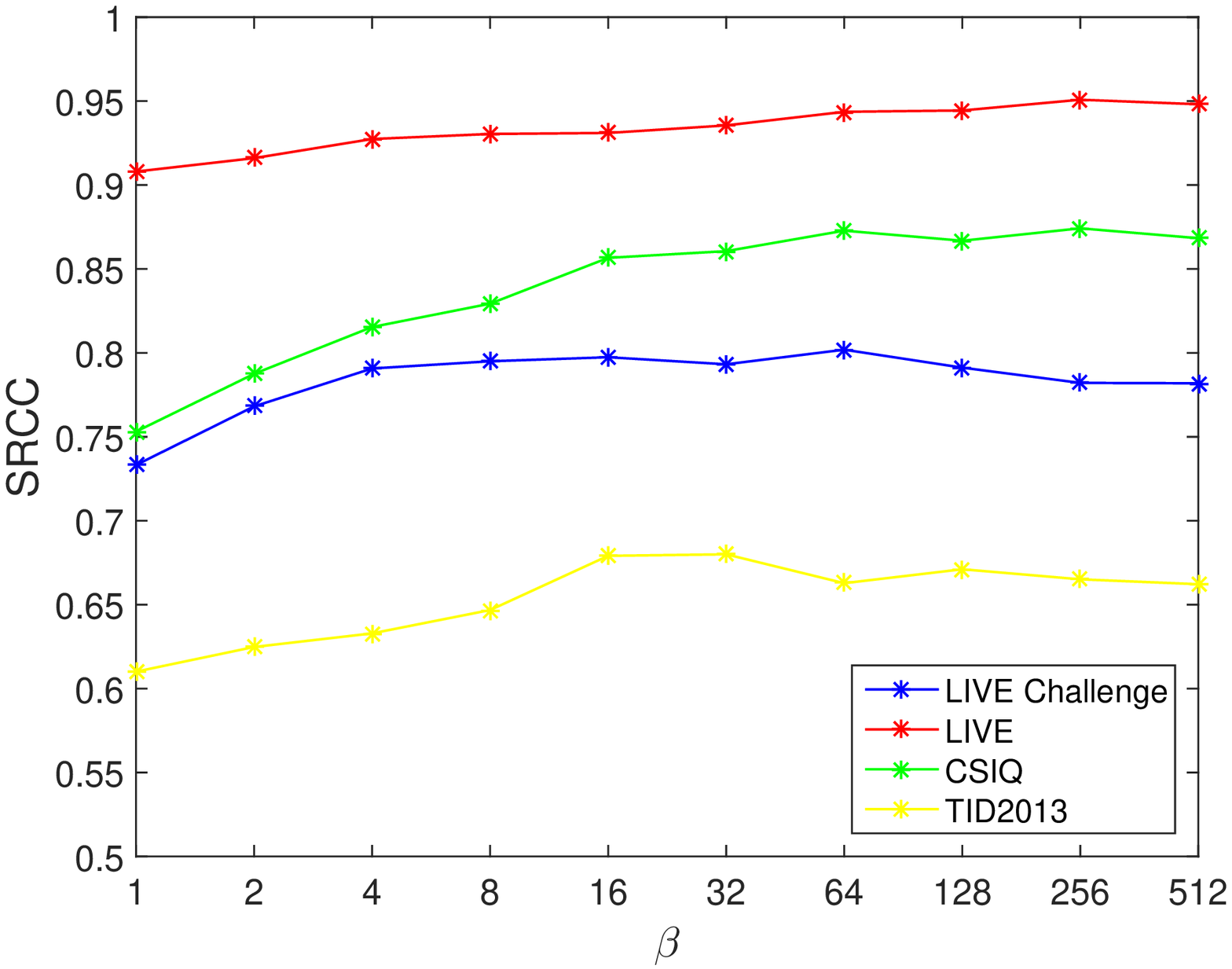}
\end{minipage}}%
\subfigure{
\begin{minipage}[t]{0.4\linewidth}
\centering
\includegraphics[width=1.0\textwidth]{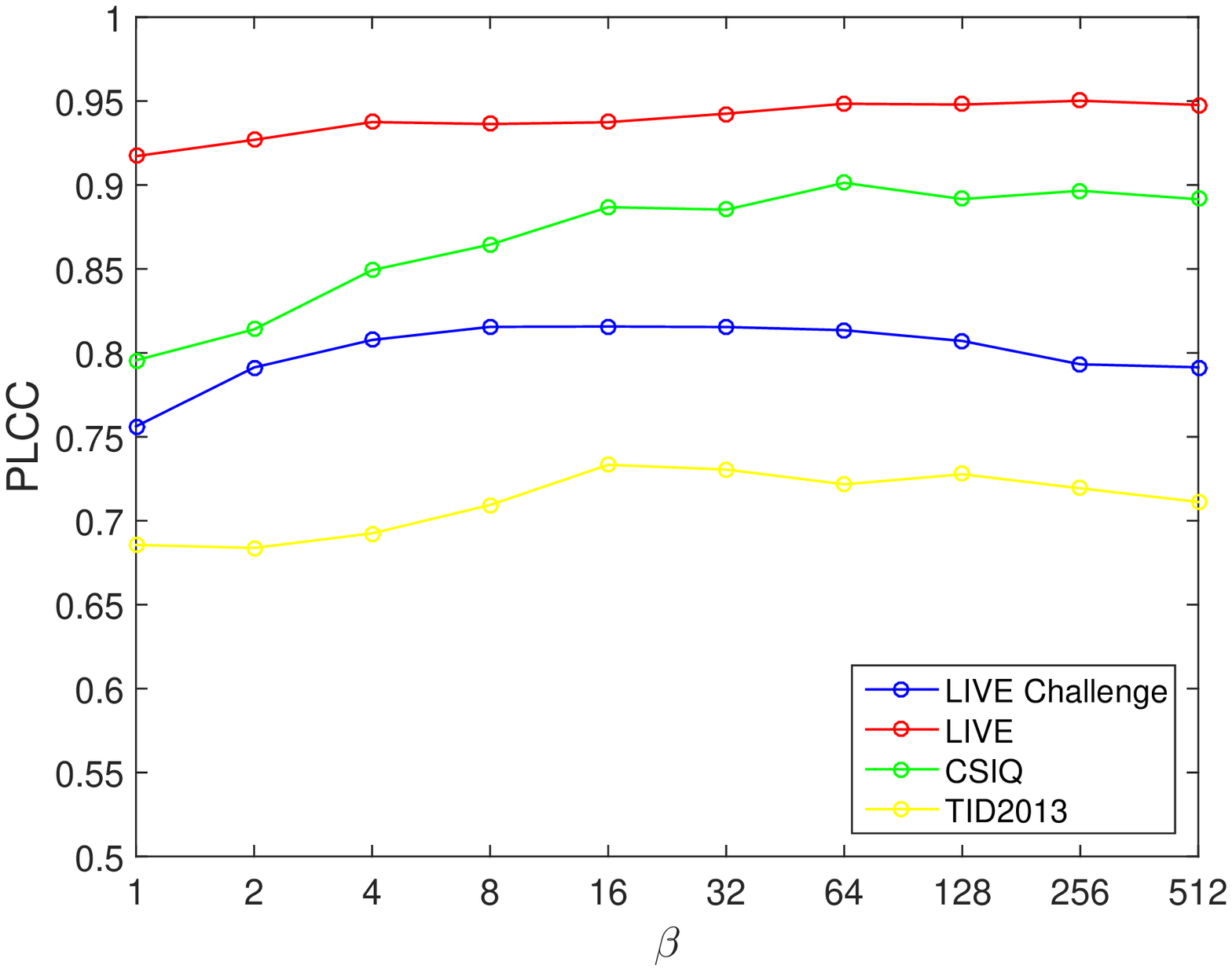}
\end{minipage}}
\caption{Plots of quality prediction performance (left: SRCC, right: PLCC) of the PQR-based deep BIQA model against the smoothing parameter $\beta$. The median SRCC and PLCC over 10 repetitions are reported on each database, using 60\% of the images for training and 20\% for validation. }
\label{figure:beta}
\end{figure*}

\begin{figure*}[t]
\centering
\subfigure[]{
\begin{minipage}[t]{0.4\linewidth}
\centering
\includegraphics[width=1.0\textwidth]{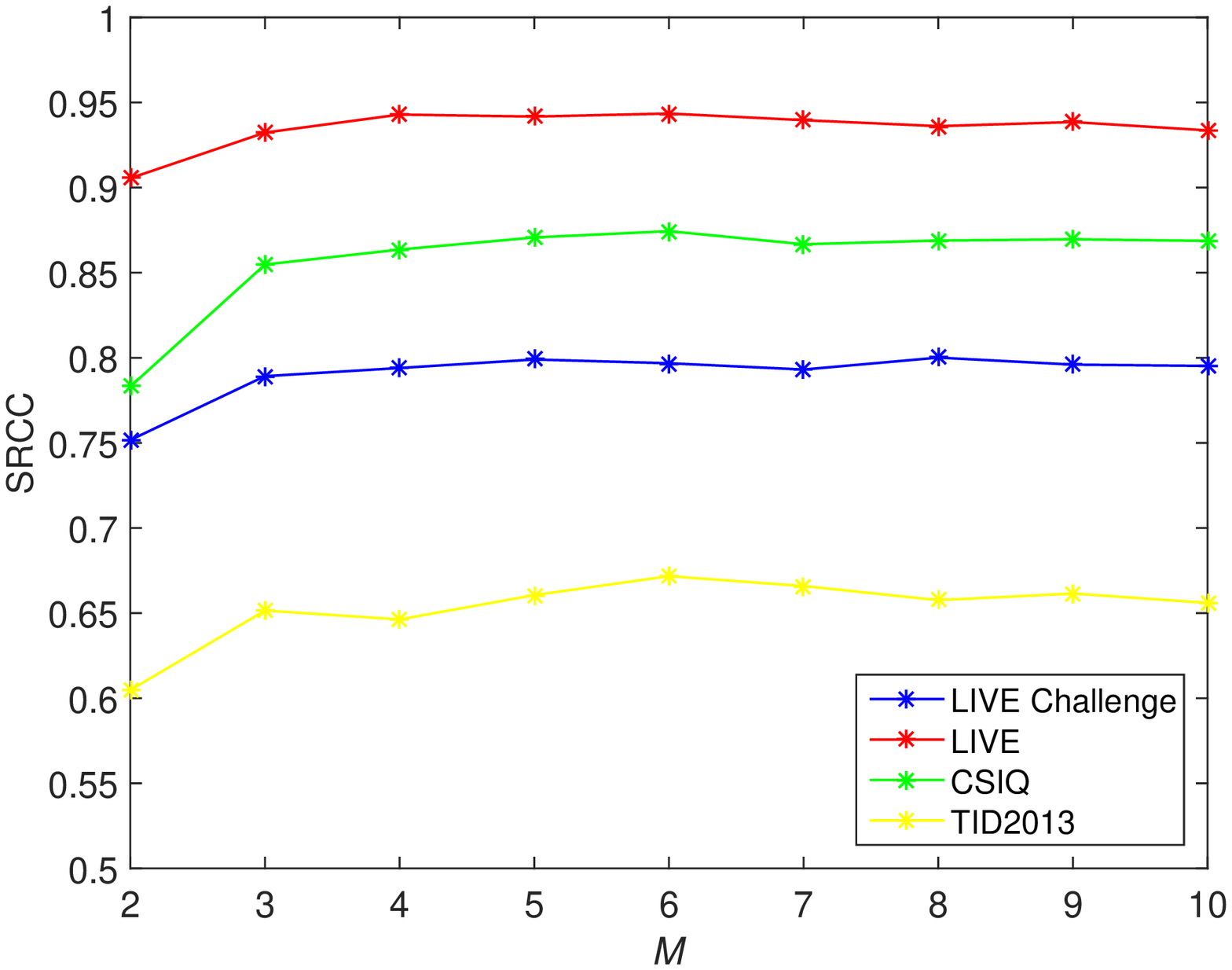}
\end{minipage}}%
\subfigure[]{
\begin{minipage}[t]{0.4\linewidth}
\centering
\includegraphics[width=1.0\textwidth]{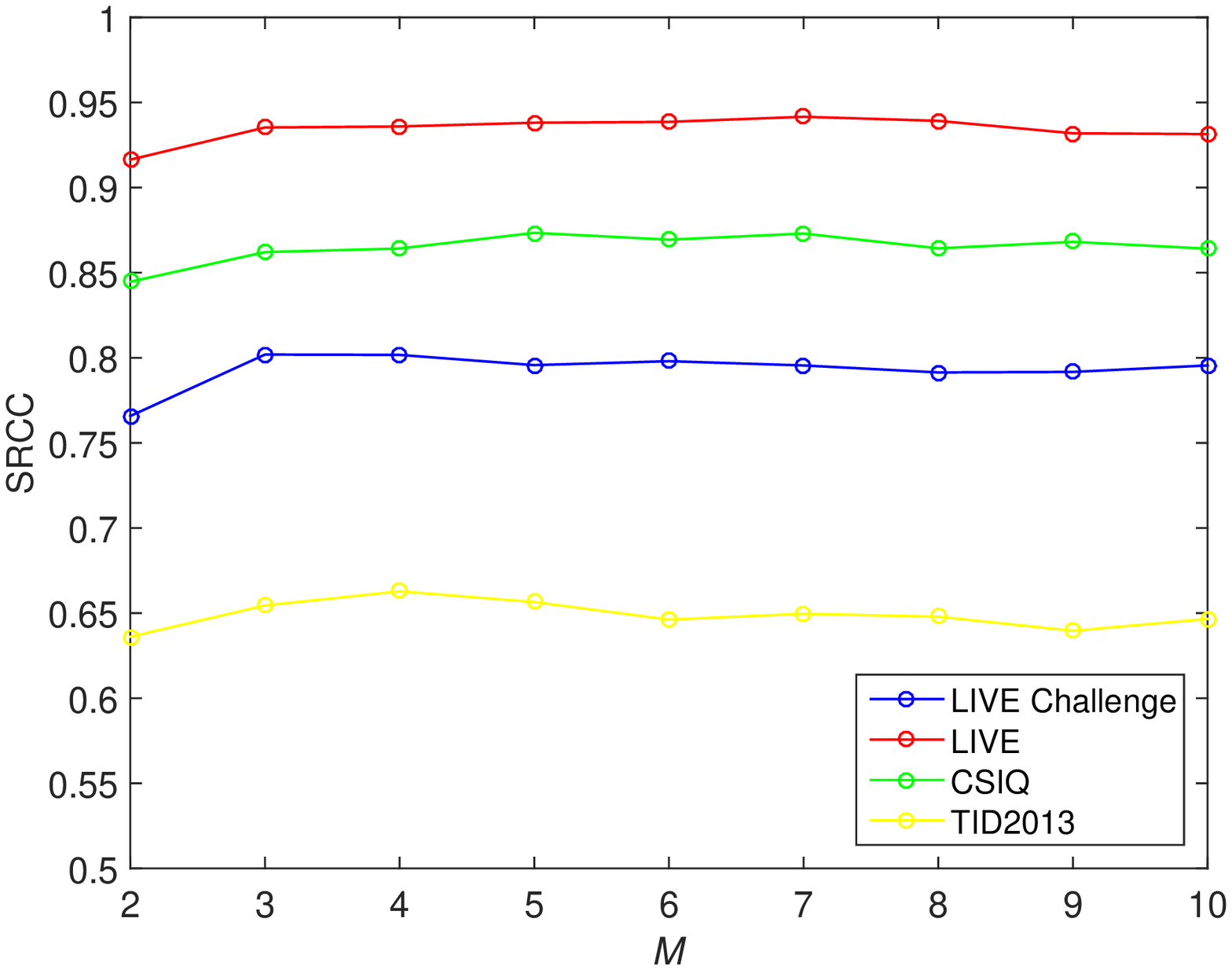}
\end{minipage}}
\caption{Plots of quality prediction performance (SRCC) of the PQR-based deep BIQA model against quality anchor quantization density parameter $M$. (a) uniform quantization; (b) Lloyd-Max quantization. The median SRCC over 10 repetitions is reported on each database, using 60\% of images for training and 20\% for validation. }
\label{figure:levels}
\end{figure*}

We first tested the effects of varying the smoothing parameter $\beta$. We evenly partitioned the quality range to determine the quality anchors and fixed $M=5$. Ten different choices of $\beta$ were tested on a log scale: $\beta = 2^{s}, s=0,1,...,9$, and evaluated on the validation set, with the results shown in Fig. \ref{figure:beta}.
As can be seen, the choice of $\beta$ substantially affects the final quality prediction performance. As $\beta$ is increased, both the SRCC and PLCC rise, then plateau and then usually decline.
Setting $\beta$ to a small value results in poor accuracy, because small values of $\beta$ cause the probabilities to become evenly distributed, making it difficult to distinguish perceptual quality levels. Choosing a very large $\beta$ would eventually cause each image to have only one none-zero quality level.
The best choice of $\beta$ varies slightly over the different databases, but $\beta = 64$ delivers uniformly excellent performance and is consequently used in all the following experiments.

We then evaluated the influence of the quality anchor quantization parameter $M$, for both uniform and Lloyd-Max quantization. We computed the achieved SRCC and PLCC of the PQR-based deep BIQA model for all integer values of $M$ in [2,10], and reported the results in the plots in Fig. \ref{figure:levels}. To conserve space, we only plotted the SRCC curves, since the PLCC curves exhibit a very similar trend. From Fig. \ref{figure:levels} we make the following observations. First, uniform quantization yields performance in parallel with Lloyd-Max except at very low densities, where both approaches suffered ($M=2, 3$). Second, the performance is stable for $M$ within [4,10], although the best choice of $M$ varies slightly with database. These results show that the PQR-based deep BIQA method is highly robust to the choice and density of quality anchors. Thus, in all the following experiments, we used uniform quantization with $M = 5$  quality anchors.

\subsection{Comparison Against Scalar Regression}

\begin{figure*}[tp]
\centering
\subfigure{
\begin{minipage}[t]{0.3\linewidth}
\includegraphics[width=1.0\textwidth]{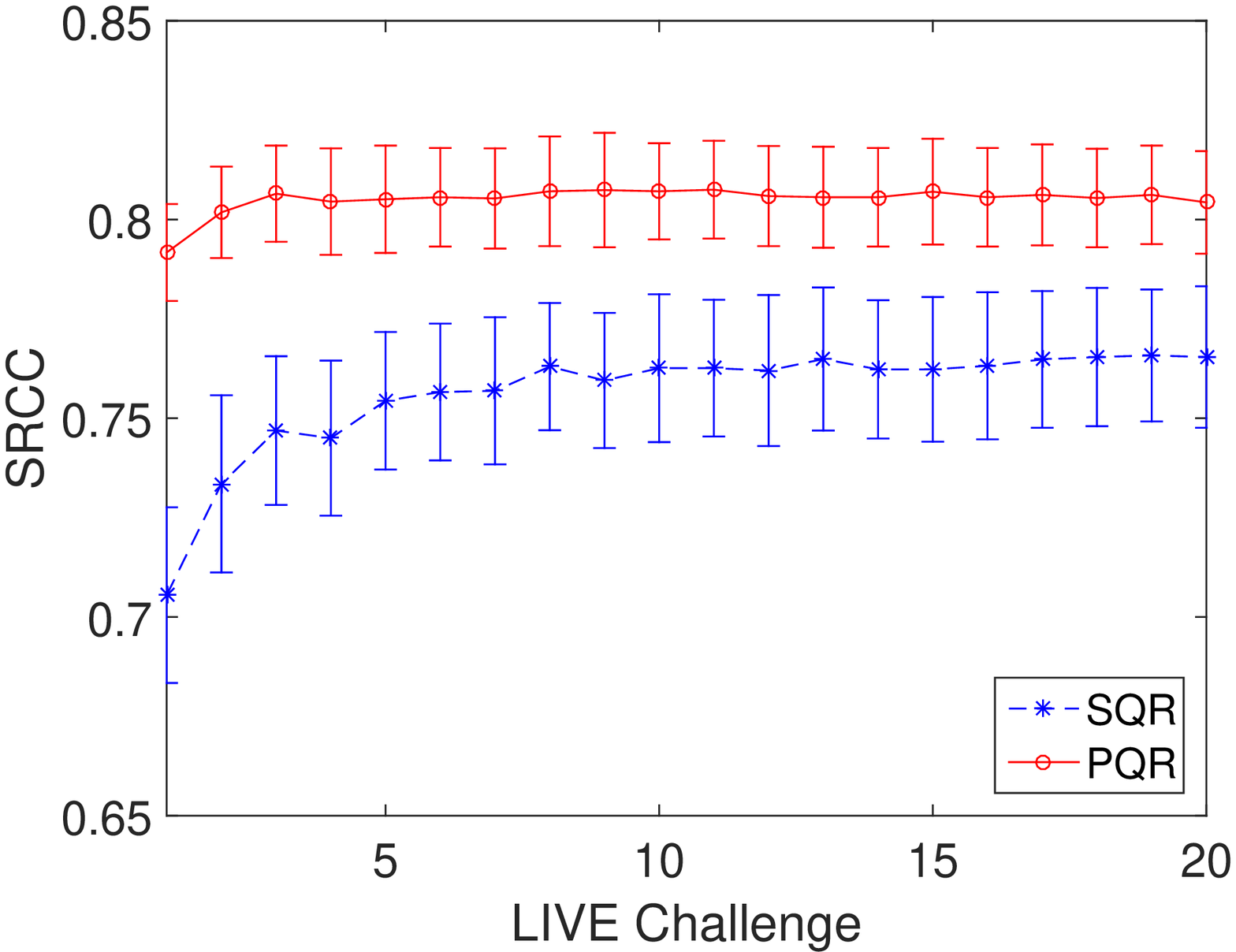}
\end{minipage}}%
\subfigure{
\begin{minipage}[t]{0.3\linewidth}
\includegraphics[width=1.0\textwidth]{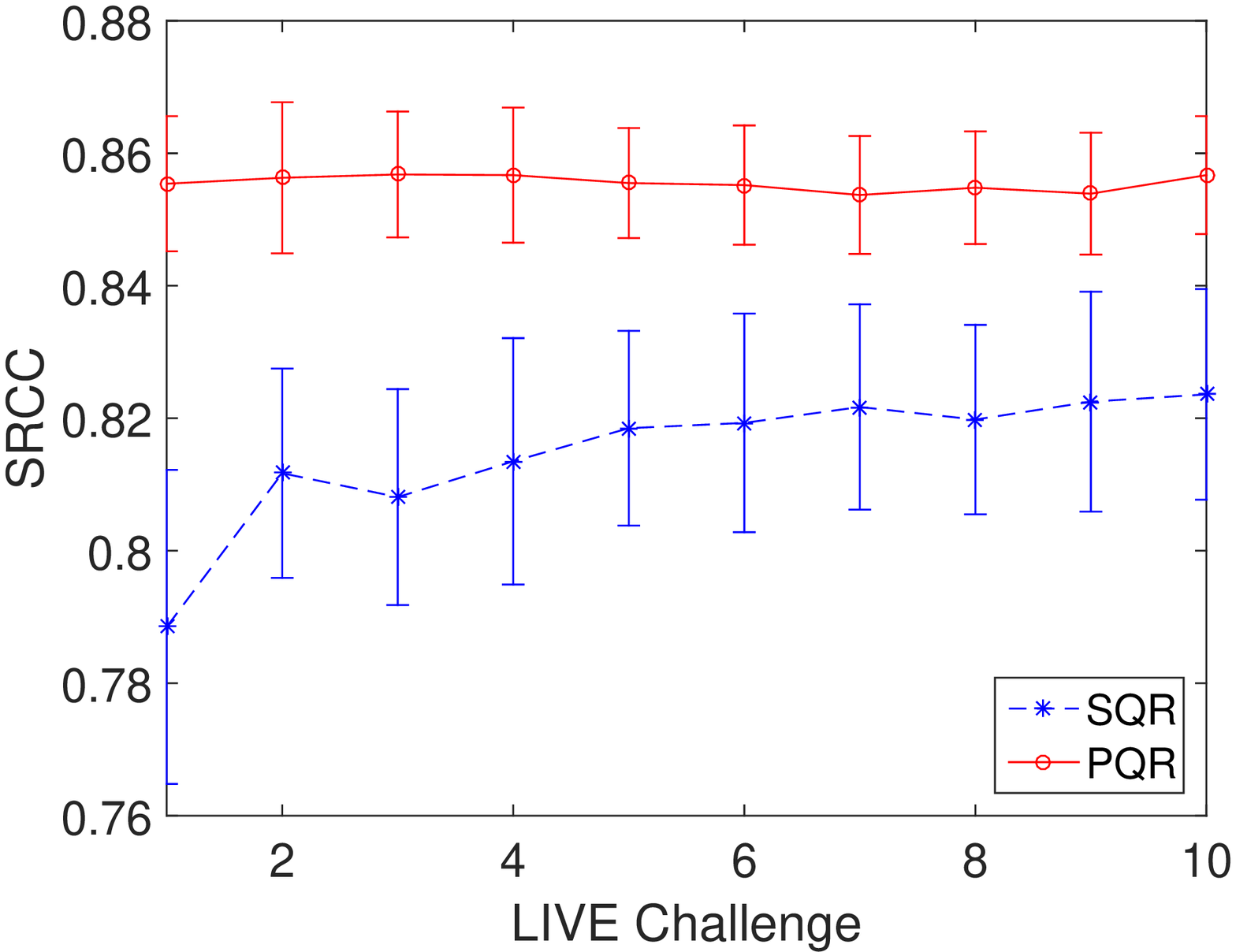}
\end{minipage}}%
\subfigure{
\begin{minipage}[t]{0.3\linewidth}
\includegraphics[width=1.0\textwidth]{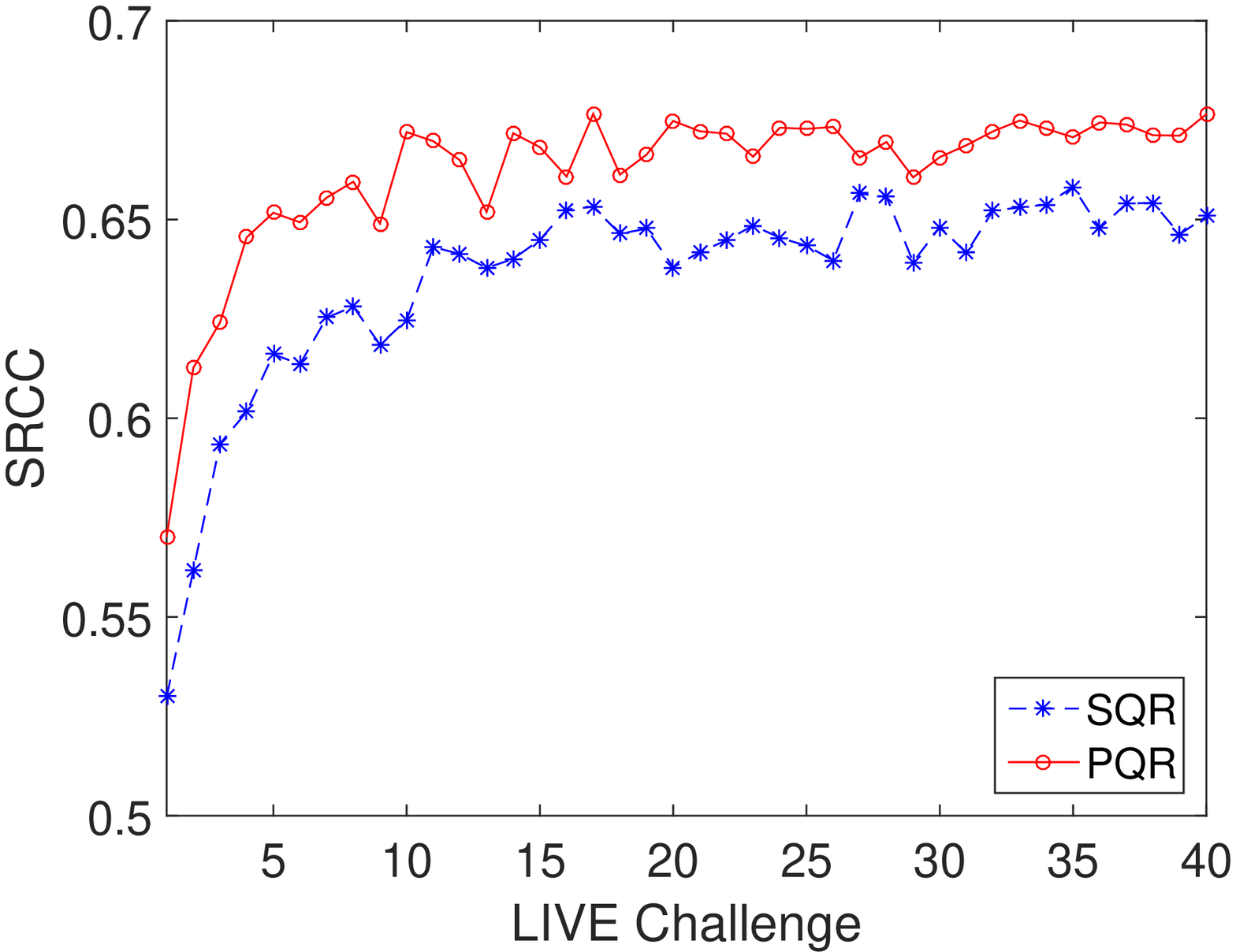}
\end{minipage}}

\subfigure{
\begin{minipage}[t]{0.3\linewidth}
\includegraphics[width=1.0\textwidth]{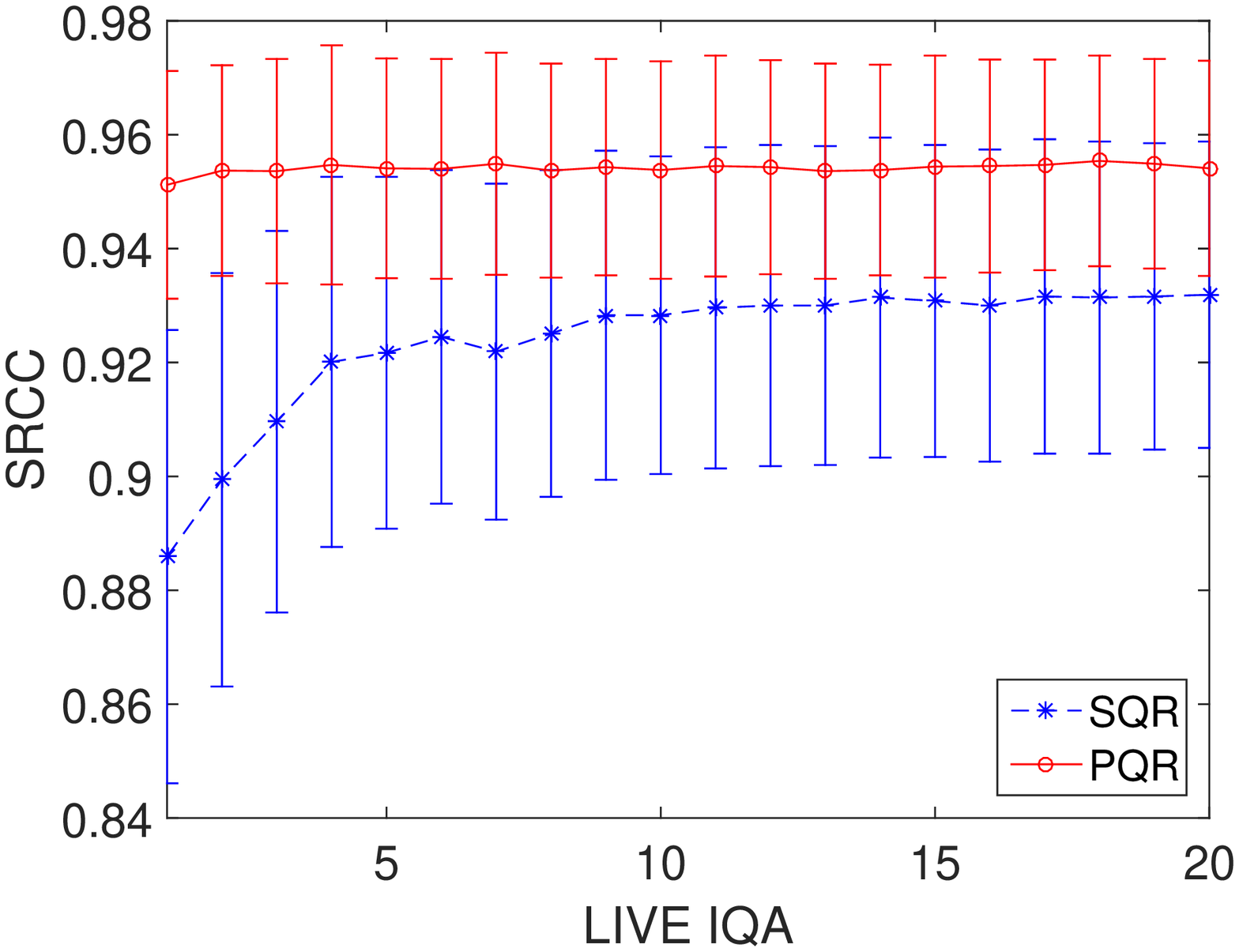}
\end{minipage}}%
\subfigure{
\begin{minipage}[t]{0.3\linewidth}
\includegraphics[width=1.0\textwidth]{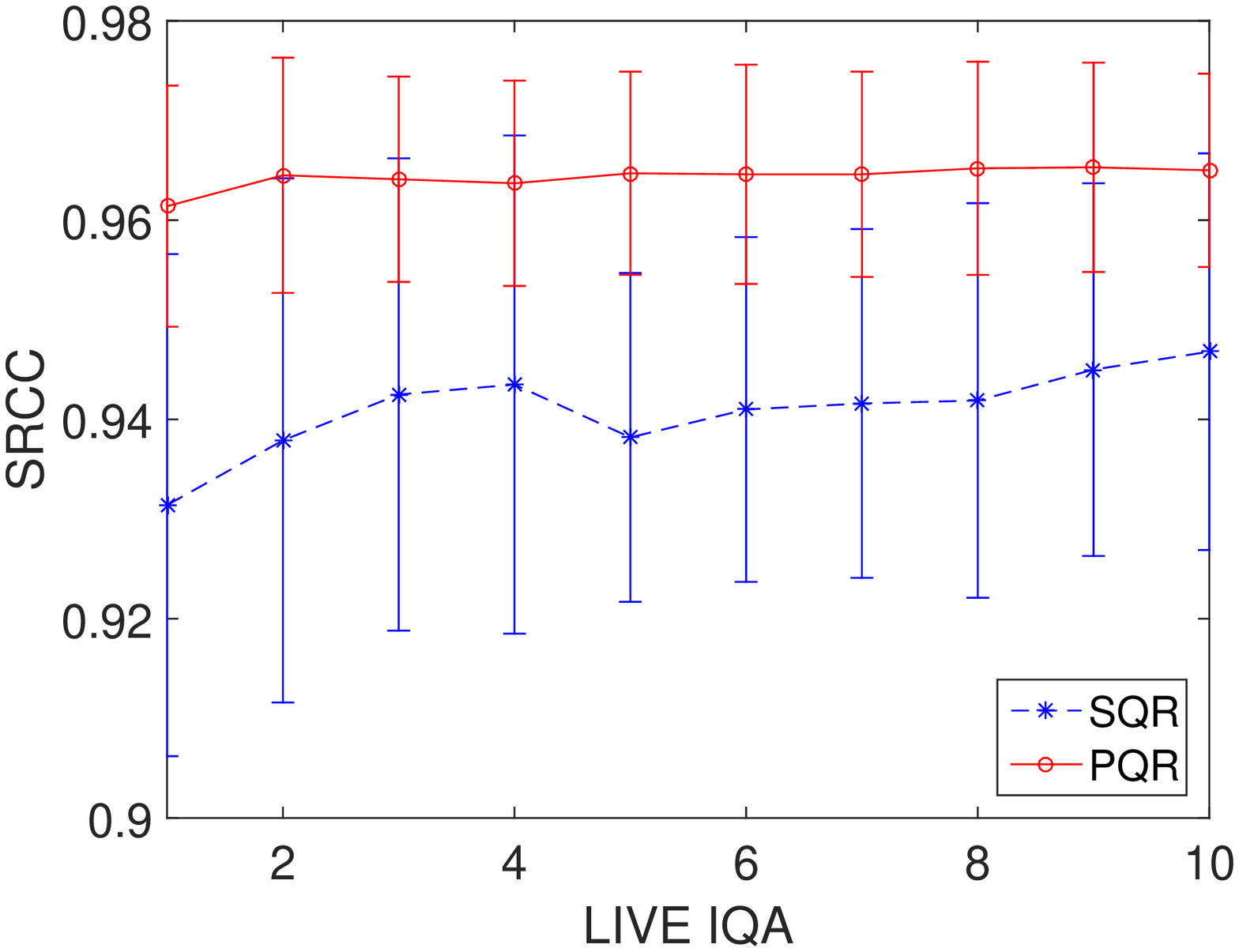}
\end{minipage}}%
\subfigure{
\begin{minipage}[t]{0.3\linewidth}
\includegraphics[width=1.0\textwidth]{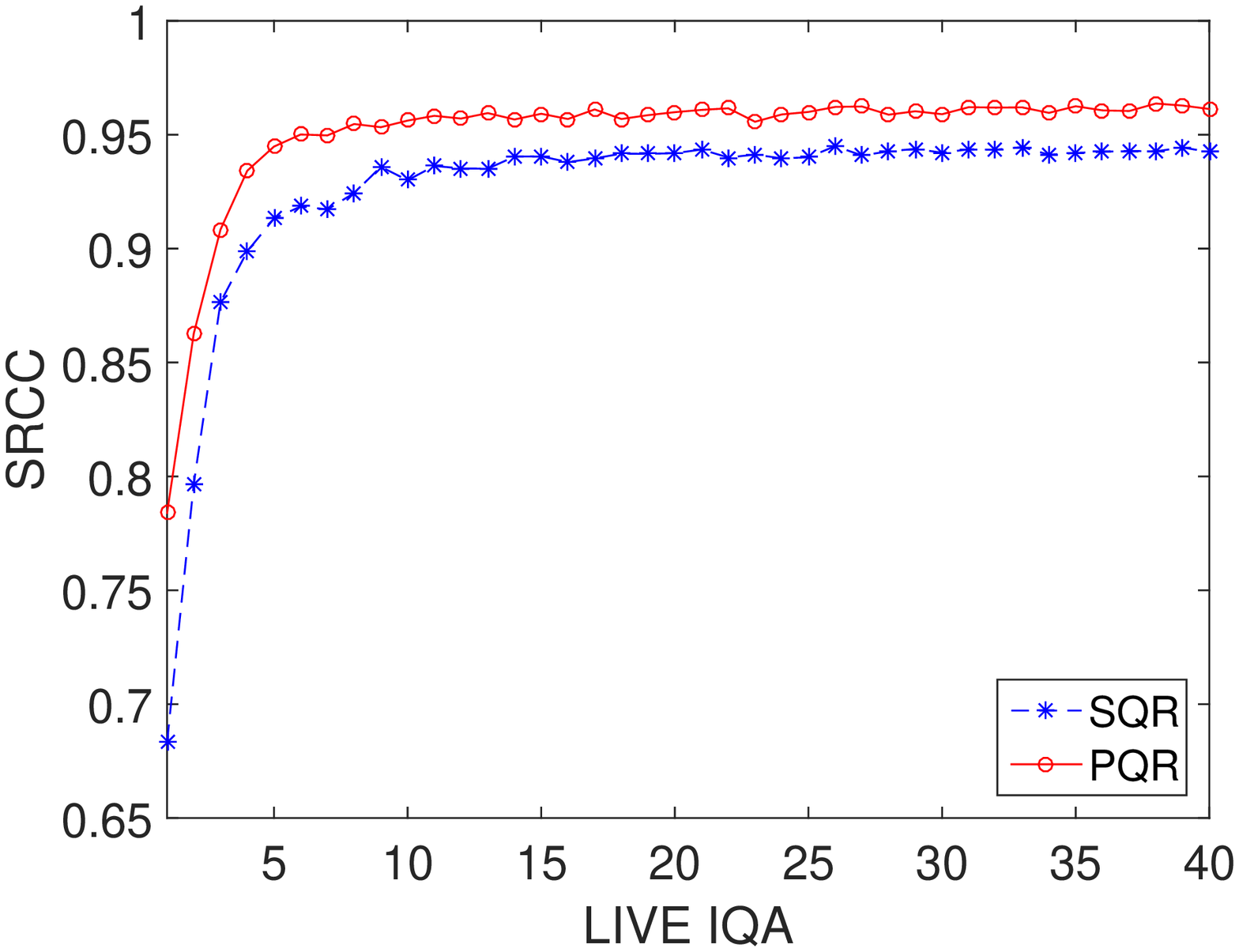}
\end{minipage}}

\subfigure{
\begin{minipage}[t]{0.3\linewidth}
\includegraphics[width=1.0\textwidth]{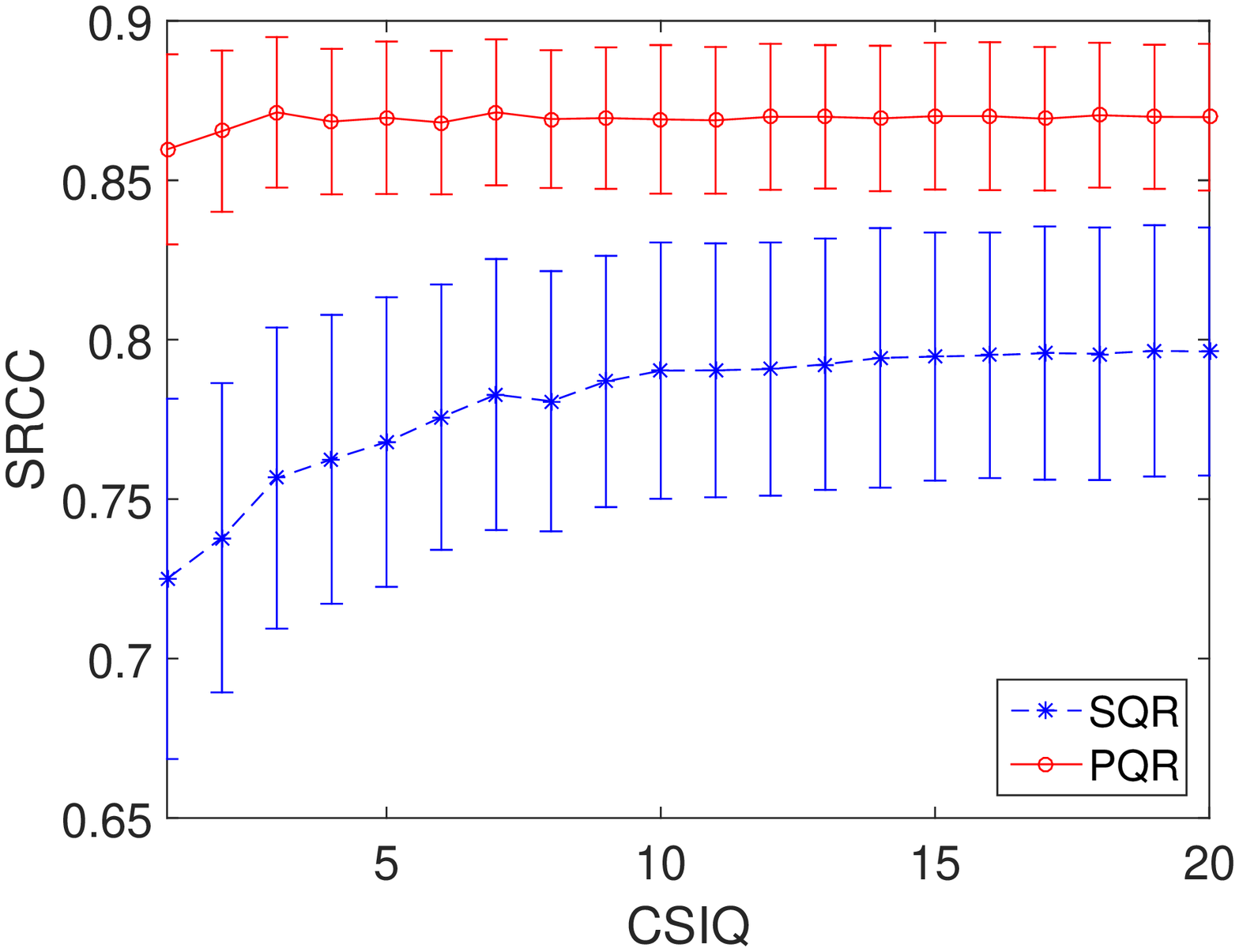}
\end{minipage}}%
\subfigure{
\begin{minipage}[t]{0.3\linewidth}
\includegraphics[width=1.0\textwidth]{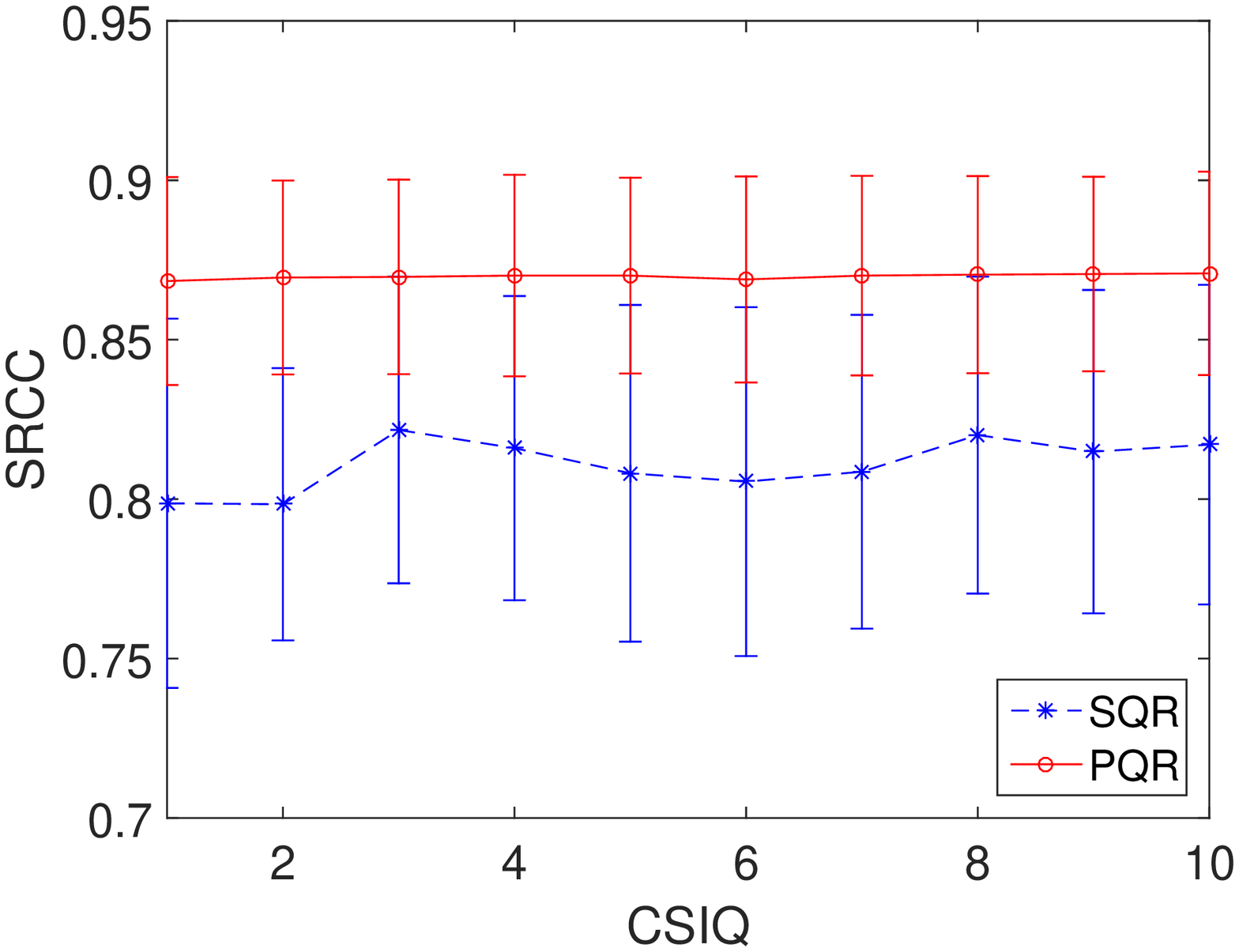}
\end{minipage}}%
\subfigure{
\begin{minipage}[t]{0.3\linewidth}
\includegraphics[width=1.0\textwidth]{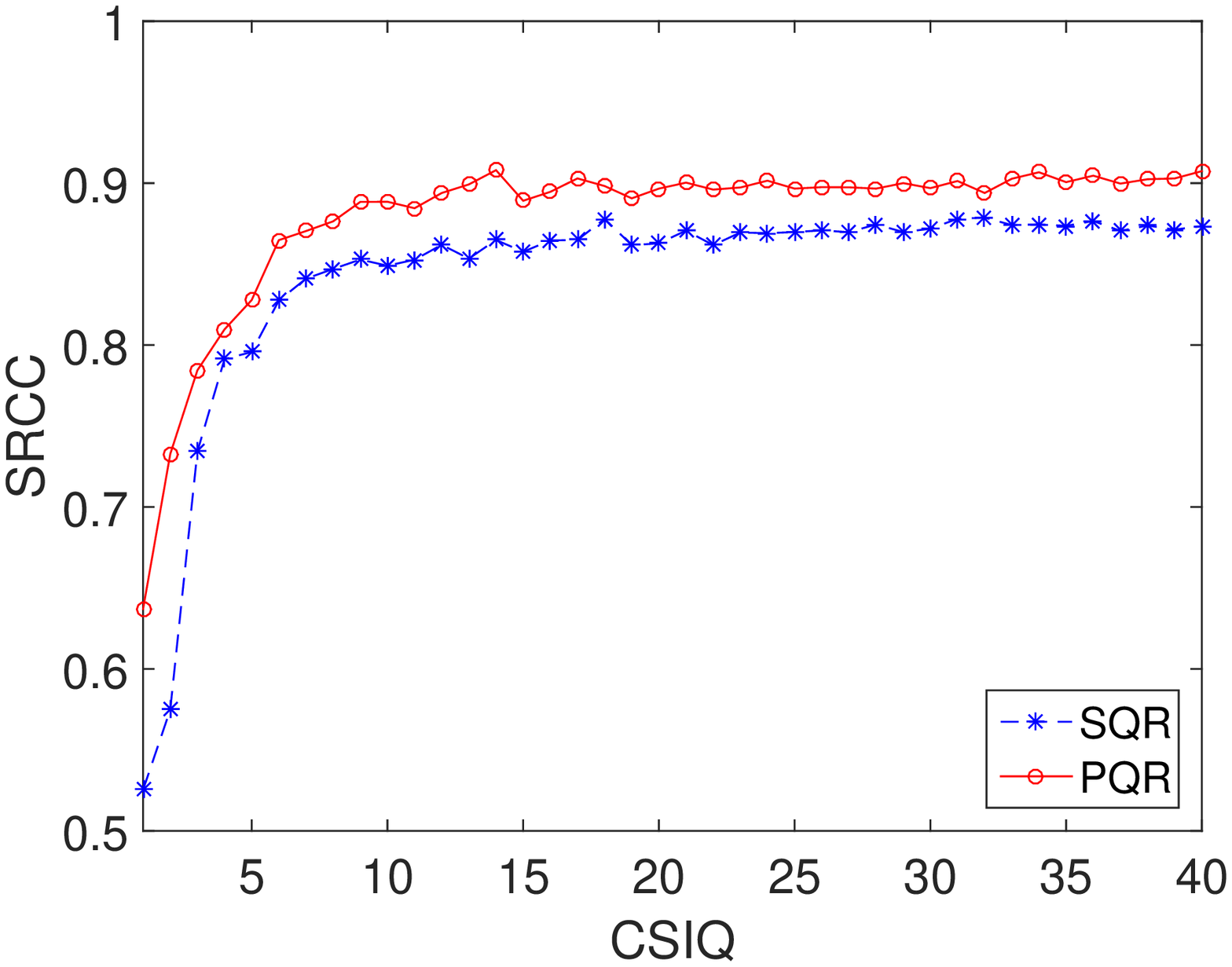}
\end{minipage}}

\subfigure{
\begin{minipage}[t]{0.3\linewidth}
\includegraphics[width=1.0\textwidth]{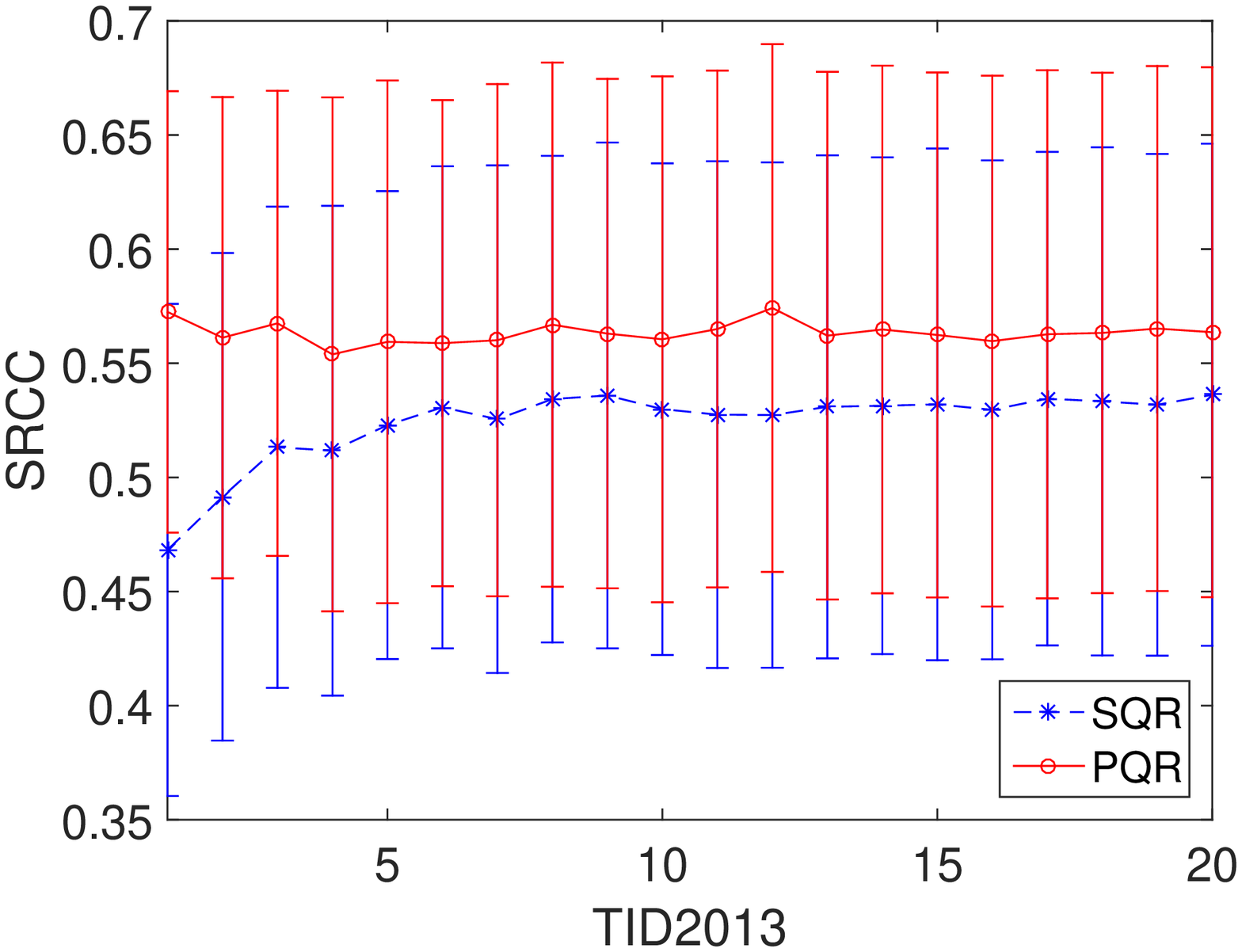}
\center\footnotesize{(a) AlexNet}
\end{minipage}}%
\subfigure{
\begin{minipage}[t]{0.3\linewidth}
\includegraphics[width=1.0\textwidth]{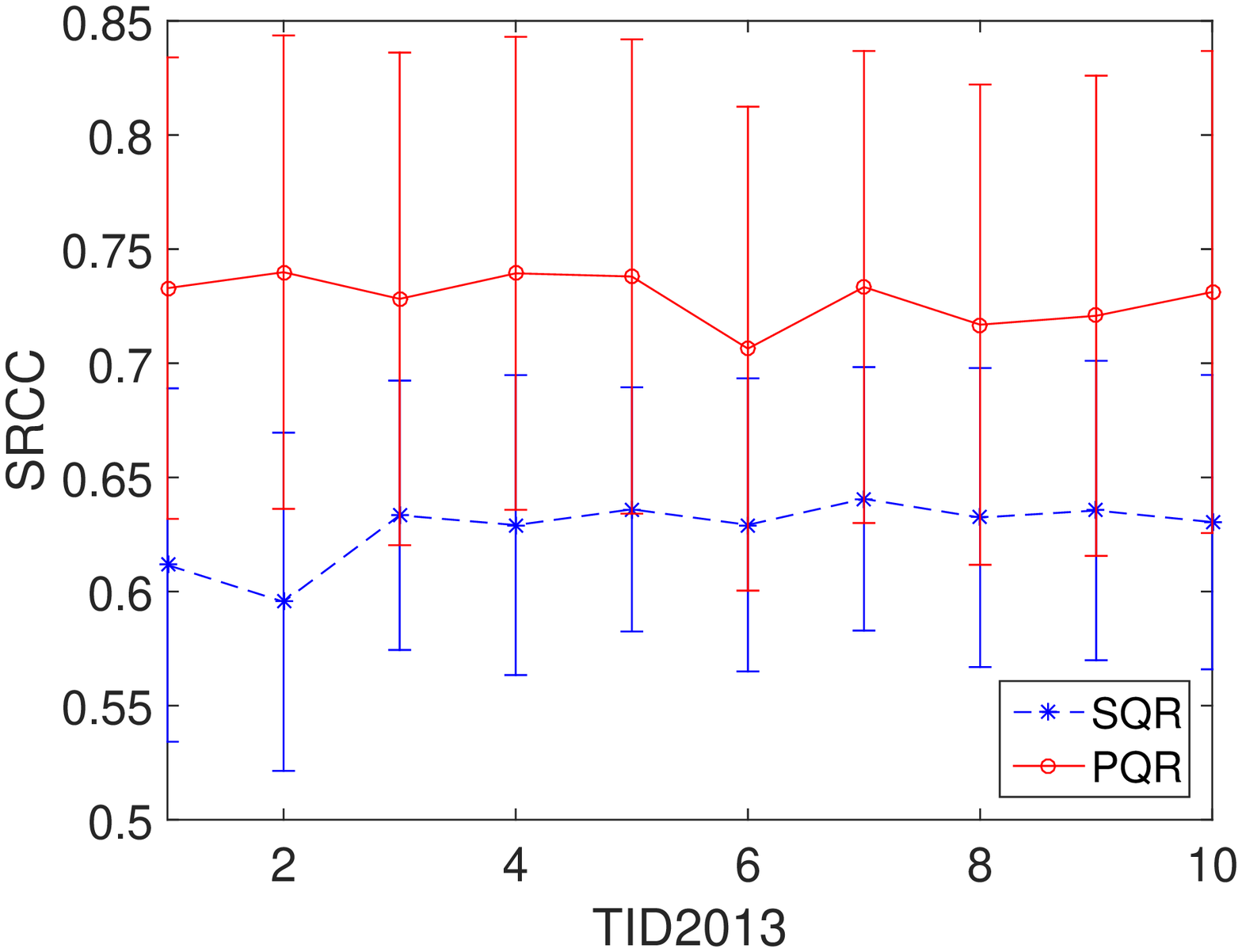}
\center\footnotesize{(b) ResNet50}
\end{minipage}}%
\subfigure{
\begin{minipage}[t]{0.3\linewidth}
\includegraphics[width=1.0\textwidth]{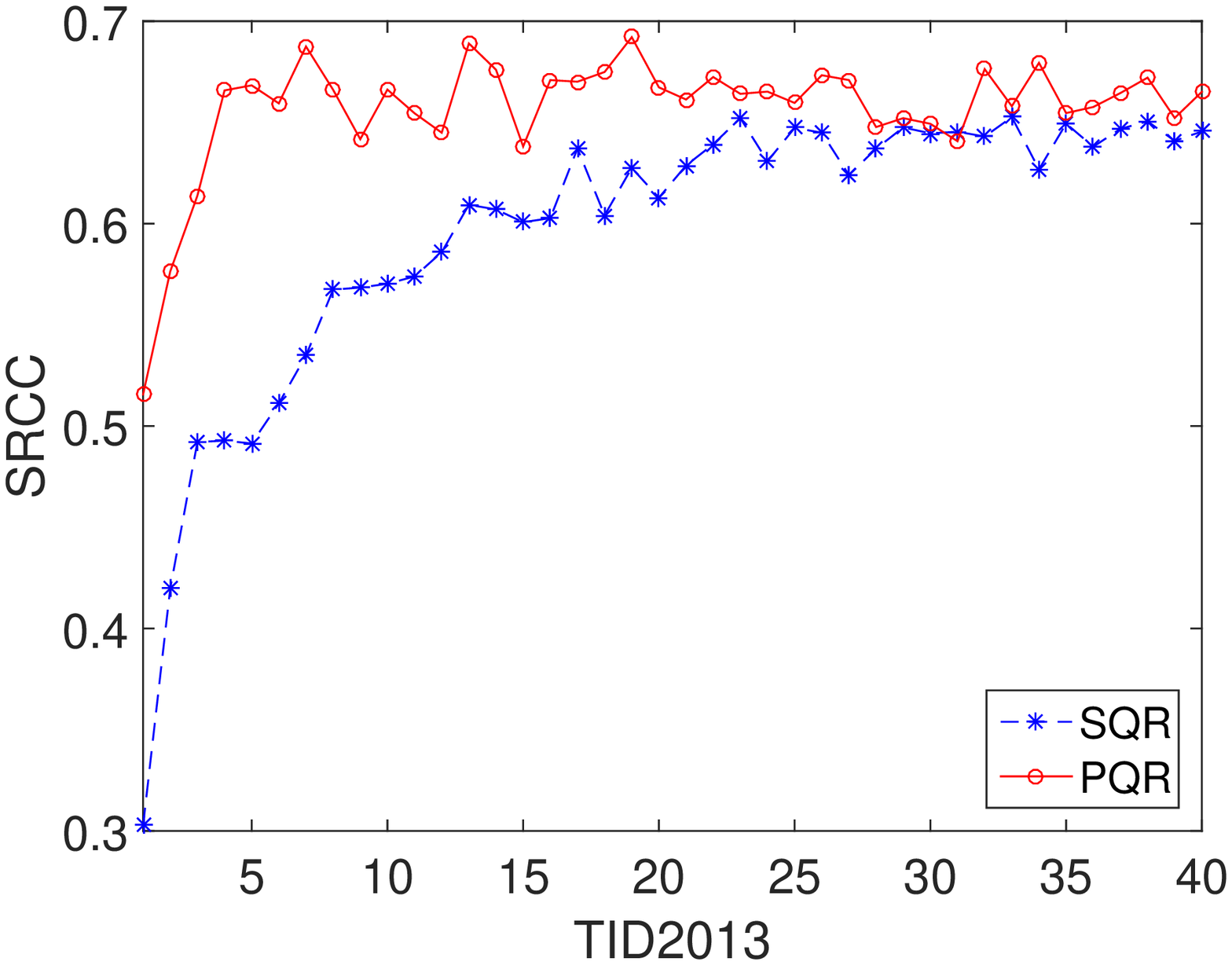}
\center\footnotesize{(c) S\_CNN}
\end{minipage}}

\caption{Performance plots that compare the proposed PQR-based model with the traditional SQR-based model over all training/fine tuning epochs. 80\% of each database's images were used for training and the remaining 20\% of the images were used for testing. The median SRCC and the corresponding standard deviation over 10 repetitions are shown for very combination of database and CNN model. }
\label{figure:convergency}
\end{figure*}

Next we compared the performances of deep BIQA models trained on PQR against commensurate models trained on traditional scalar quality representation (SQR) in regards to both convergence speed and prediction accuracy. For fair comparison, all settings except for the loss function were held constant across all models. On each database, 80\% of the images were used for training and the remaining 20\% of the images were used for testing. For the AlexNet and ResNet50 models, the median SRCC values and the corresponding standard deviations across 10 repetitions for each training epoch are shown in Figs. \ref{figure:convergency}(a) and \ref{figure:convergency}(b), respectively. Fig. \ref{figure:convergency}(c) shows the results of S\_CNN, where the standard deviations are not plotted to make the curves more distinguishable. The best median SRCC and PLCC values among all epochs are reported in Table \ref{table:predictors}. We did not compare the converged values of the two loss functions because they are in different units.

\begin{table*}[t]
\centering
\caption{Comparing the best performance of PQR against SQR. 80\% of each database's images were used for training while the remaining 20\% of the images were used for testing. The median SRCC and PLCC across 10 repetitions are reported for each combination. The Best Performances On Each Database Are Boldfaced.}
\label{table:predictors}
{\renewcommand{\arraystretch}{1.0}
\begin{tabular}{|=c|+c|+c+c|+c+c|+c+c|+c+c|}
\hline
\multirow{2}{*}{CNN model} & \multirow{2}{*}{Methods} & \multicolumn{2}{c|}{LIVE Challenge}  & \multicolumn{2}{c|}{LIVE IQA} & \multicolumn{2}{c|}{CSIQ} & \multicolumn{2}{c|}{TID2013}    \\\cline{3-10}
                &   & SRCC  & PLCC    & SRCC  & PLCC   & SRCC  & PLCC  & SRCC  & PLCC       \\\hline
\multirow {2}{*}{AlexNet}
                        & SQR          & 0.7658 & 0.8074 & 0.9319 & 0.9462 & 0.7965 & 0.8405 & 0.5362 & 0.6136      \\
                        & PQR        & 0.8075 & 0.8357 & 0.9554 & 0.9638 & 0.8713 & 0.8958 & 0.5742 & 0.6687      \\\hline
\multirow {2}{*}{ResNet50}
                        & SQR          & 0.8236 & 0.8680 & 0.9468 & 0.9527 & 0.8217 & 0.8713 & 0.6406 & 0.7068     \\
                        & PQR        & \textbf{0.8568} & \textbf{0.8822} & \textbf{0.9653} & \textbf{0.9714} & 0.8728 & 0.9010 & \textbf{0.7399} & \textbf{0.7980}      \\\hline
\multirow {2}{*}{S\_CNN}
                        & SQR          & 0.6582 & 0.6729 & 0.9450 & 0.9455 & 0.8787 & 0.8987 & 0.6526 & 0.6921      \\
                        & PQR        & 0.6766 & 0.7032 & 0.9637 & 0.9656 & \textbf{0.9080} & \textbf{0.9267} & 0.6921 & 0.7497        \\\hline
\end{tabular}}
\end{table*}

We can immediately make the following observations regarding the results shown in Fig. \ref{figure:convergency} and Table \ref{table:predictors}. First, the BIQA models trained using our proposed PQR model significantly outperform the BIQA models trained by the traditional SQR model on all the three CNN architectures. This convincingly shows that our proposed PQR model is a very effective new tool for deep BIQA model learning.
Secondly, under the same settings, the BIQA model training using the PQR model converges much faster than the SQR model, especially in regards to fine tuning the deep models, where we found that the PQR-based model converges within no more than 3 epochs on all of the databases.\footnote{The fast convergence of our PQR model was not caused by the learning rate, which was the same for both the PQR and SQR models. Indeed, the mean-squared loss function used in the SQR model is much more fragile than the softmax cross-entropy loss used in the PQR model. We have observed that increasing the learning rate for the SQR model can cause the training process to oscillate or even fail to converge, whereas our PQR model can handle a much wider range of learning rates.}
Finally, our method results in a much smaller standard deviation of prediction performance on most of the databases (except for TID2013), which strongly suggests that the probabilistic representation is much more robust and stable than directly regressing the deterministic scalar quality scores using the mean squared error loss. All of the trained BIQA models have standard deviations of performance exceeding 0.1 on the TID2013 database (e.g., the SRCC of ResNet50 using PQR varies from 0.6 to 0.9 over the 10 repetitions). This behavior may be caused by peculiarities of the database, which contains a variety of rare or unrealistic distortions.

\subsection{Comparison Among Different CNN Models}

A very interesting observation can be made on Table \ref{table:predictors}: that the S\_CNN trained from scratch can achieve performance that is competitive, or even superior to that achieved by the two pre-trained deep CNN models (AlexNet and ResNet50) on the three legacy databases, but much worse than the two deep models on the LIVE Challenge.

This divergence in performance can likely be explained in terms of the different characteristics of the databases.
The legacy databases (notably LIVE IQA and CSIQ) contain a limited variety of synthetic distortion types and degradation levels, which have been homogeneously applied in isolation to a small number of source images. Because of this, the mapping from perceptual quality degradation to quality scores is relatively easy to learn, even by a shallow CNN model. Moreover, the greater degree of spatial distortion homogeneity makes it possible to leverage small image patches when training a shallow CNN model, because they are more representative of the distortions afflicting the whole image. Several previous methods have demonstrated the effectiveness of using patch-wise training as a method of data augmentation on the legacy synthetic distortion databases \cite{kang2014convolutional,tang2014blind,hou2015blind,kim2017fully}. However, the LIVE Challenge and TID2013 databases are both very difficult. LIVE Challenge contains many highly diverse contents that are authentically distorted in many complex combinations and degrees. These complex, real-world multi-distortions are often quite inhomogeneous. LIVE Challenge is simply too complex for a shallow CNN model to be able to achieve good performance. Blindly assessing the quality of real-world distortions is a very difficult problem that appears to require either much more data or innovations in network design.

Deep CNN models pre-trained on target problem like the ImageNet classification task can generalize well to other image recognition and processing tasks \cite{donahue2014decaf,sharif2014cnn}. However, the transferability of a network depends on the degrees of statistical similarity between the training data and target data \cite{yosinski2014transferable}. Since both ImageNet and LIVE Challenge consist of natural images afflicted with authentic distortions, it is natural to infer that models pre-trained on ImageNet can transfer well to task on the LIVE Challenge database. By contrast, the three legacy databases are composed of images algorithmically modified by synthetic distortions, many of which are exotic, rare, or even unlikely. It follows that the statistics of the images in these databases are very different from those in ImageNet, hence the transferability of models pre-trained on ImageNet is greatly reduced, especially on TID2013.

\subsection{Comparison with Other BIQA methods}

\begin{table*}[t]
\centering
\caption{Comparisons with existing BIQA models. 80\% of the images were used for training and the remaining 20\% of the images were used for testing. The median SRCC and PLCC over 10 random repetitions are reported for each case. The Best Performances On Each Database Are Boldfaced.}
\label{table:other methods}
\begin{tabular}{|c|cc|cc|cc|cc|}
\hline
\multirow {2}{*}{Methods} &\multicolumn{2}{c|}{LIVE Challenge}  & \multicolumn{2}{c|}{LIVE IQA} & \multicolumn{2}{c|}{CSIQ} & \multicolumn{2}{c|}{TID2013} \\\cline{2-9}
                                            &  SRCC  & PLCC    & SRCC  & PLCC   & SRCC  & PLCC  & SRCC  & PLCC        \\\hline
DIIVINE \cite{moorthy2011blind}             &0.58 $\pm$ 0.03 & 0.60 $\pm$ 0.03 & 0.8787 & 0.8813 & 0.7835 & 0.8362 & 0.5829 & 0.6723 \\
CORNIA \cite{ye2012unsupervised}            &0.63 $\pm$ 0.04 & 0.66 $\pm$ 0.04 & 0.9420 & 0.9457 & 0.7299 & 0.8036 & 0.6226 & 0.7038 \\
BRISQUE \cite{mittal2012no}                 &0.61 $\pm$ 0.03 & 0.65 $\pm$ 0.04 & 0.9374 & 0.9448 & 0.7502 & 0.8286 & 0.5258 & 0.6331 \\
NIQE \cite{mittal2013making}                &0.43 $\pm$ 0.03 & 0.48 $\pm$ 0.03 & 0.9154 & 0.9194 & 0.6298 & 0.7181 & 0.2992 & 0.4154 \\
IL-NIQE \cite{zhang2015feature}             &0.43 $\pm$ 0.03 & 0.51 $\pm$ 0.02 & 0.9017 & 0.8654 & 0.8066 & 0.8083 & 0.5185 & 0.6398 \\
HOSA  \cite{hou2015blind}                   &0.66 $\pm$ 0.04 & 0.68 $\pm$ 0.03 & 0.9477 & 0.9492 & 0.7812 & 0.8415 & 0.6876 & 0.7637 \\
FRIQUEE-ALL \cite{ghadiyaram2016perceptual} &0.69 $\pm$ 0.03 & 0.71 $\pm$ 0.03 & 0.9507 & 0.9576 & 0.8414 & 0.8733 & 0.7133 & 0.7755 \\
\hline
PQR (AlexNet)                              &0.81 $\pm$ 0.01 & 0.84 $\pm$ 0.01 & 0.9554 & 0.9638 & 0.8713 & 0.8958 & 0.5743 & 0.6687 \\
PQR (ResNet50)                             &\textbf{0.86 $\pm$ 0.01} & \textbf{0.88 $\pm$ 0.01} &\textbf{0.9653} & \textbf{0.9714} & 0.8728 & 0.9010 & \textbf{0.7399} & \textbf{0.7980} \\
PQR (S\_CNN)                               &0.68 $\pm$ 0.03 & 0.70 $\pm$ 0.03 & 0.9637 & 0.9656 & \textbf{0.9080} & \textbf{0.9267}& 0.6921 & 0.7497 \\
\hline\hline
Kang \textit{et al.} \cite{tang2014blind}   & N.A. & N.A. & 0.956 & 0.953 & N.A. & N.A.  & N.A. & N.A.          \\
Hou \textit{et al.} \cite{hou2015blind}     & N.A. & N.A. & 0.930  & 0.927  & N.A. & N.A. & N.A. & N.A.            \\
\hline
\end{tabular}
\end{table*}

\begin{table*}[t]
\centering
\caption{Cross database evaluation. Each entire database was used for both training and testing. Only SRCC is reported.}
\label{table:cross database}
{\renewcommand{\arraystretch}{1.0}
\begin{tabular}{|=c|+c|+c+c+c|+c+c+c|}
\hline
\multicolumn{2}{|c|}{Training database}  & \multicolumn{3}{c|}{LIVE Challenge}  & \multicolumn{3}{c|}{LIVE IQA} \\\hline
\multicolumn{2}{|c|}{Testing database} & LIVE IQA  & CSIQ & TID2013  & LIVE Challenge  & CSIQ   & TID2013  \\\hline
\multirow {2}{*}{AlexNet}
                        & SQR          & 0.4917 & 0.4833 & 0.2524 & 0.4972 & 0.6453 & 0.6007       \\
                        & PQR        & 0.5248 & 0.5433 & 0.3290 & 0.5498 & 0.7275 & 0.5759      \\\hline
\multirow {2}{*}{ResNet50}
                        & SQR          & 0.4158 & 0.4716 & 0.3037 & 0.5622 & 0.7188 & 0.6185     \\
                        & PQR        & 0.4396 & 0.5375 & 0.3374 & 0.5470 & 0.7169 & 0.5512      \\\hline
\multirow {2}{*}{S\_CNN}
                        & SQR          & 0.2891 & 0.3494 & 0.1520 & 0.4408 & 0.6928 & 0.4398    \\
                        & PQR        & 0.3872 & 0.3705 & 0.1917 & 0.3980 & 0.6836 & 0.4315        \\\hline
\hline
\multicolumn{2}{|c|}{Training database}  & \multicolumn{3}{c|}{CSIQ} & \multicolumn{3}{c|}{TID2013}    \\\hline
\multicolumn{2}{|c|}{Testing database} & LIVE Challenge  & LIVE IQA  & TID2013 & LIVE Challenge & LIVE IQA & CSIQ       \\\hline
\multirow {2}{*}{AlexNet}
                        & SQR          & 0.3793 & 0.8497 & 0.4314 & 0.3539 & 0.8327 & 0.6627      \\
                        & PQR        & 0.4314 & 0.8945 & 0.5471 & 0.2519 & 0.8343 & 0.6422     \\\hline
\multirow {2}{*}{ResNet50}
                        & SQR          & 0.4806 & 0.9217 & 0.5652 & 0.3514 & 0.9000 & 0.6323     \\
                        & PQR        & 0.4793 & 0.9302 & 0.5462 & 0.3157 & 0.8908 & 0.6324      \\\hline
\multirow {2}{*}{S\_CNN}
                        & SQR          & 0.4251 & 0.9119 & 0.4477 & 0.4191 & 0.8733 & 0.7143    \\
                        & PQR        & 0.4888 & 0.9067 & 0.4994 & 0.3818 & 0.8458 & 0.7330      \\\hline
\end{tabular}}
\end{table*}

We also compared the proposed PQR-based model against existing BIQA methods on the four databases. Since the split of training and testing sets will affect prediction accuracy, for a fair comparison, we re-ran the source codes of the DIIVINE \cite{moorthy2011blind}, CORNIA \cite{ye2012unsupervised}, BRISQUE \cite{mittal2012no}, NIQE \cite{mittal2013making}, IL-NIQE \cite{zhang2015feature}, HOSA \cite{hou2015blind} and FRIQUEE-ALL \cite{ghadiyaram2016perceptual} models using the same training and testing splits as we used for the PQR model.\footnote{We will release the randomly generated training and testing splits for all databases as well as our implementations of all the competing models.} Since both NIQE and IL-NIQE do not require any training, we directly evaluated them on the testing sets. For the other learning based methods, we used the source codes provided by the authors to extract features, and re-trained an SVR model using the RBF kernel (except for CORNIA and HOSA, which used the linear SVR) for each split. For the two-step DIIVINE model, we skipped the first step of identifying distortion types, and instead directly trained the SVR model using the extracted features. The parameters of the SVR were optimized for each method on each database.
The median SRCC and PLCC over 10 random rounds are reported in Table \ref{table:other methods}. To save space, we only report the standard deviations of the competing methods on the LIVE Challenge database. For those methods whose source codes are not provided by the authors, or could not be readily used to re-train the model, we simply reported the results provided in the original papers.

From Table \ref{table:other methods}, it can be observed that our probabilistic deep BIQA model achieves standout results on the LIVE Challenge database. When using ResNet50, it outperforms the previous best results (obtained by FRIQUEE-ALL) by more than 0.15 in regards to both SRCC and PLCC. This again demonstrates that fine tuning pre-trained deep models using the PQR model is an effective way to improve automatically generated predictions of the perceptual quality of images suffering from difficult authentic distortions. It is also evident that the performance of ResNet50 is better than that of AlexNet in most cases, which indicates that using more powerful pre-trained deep models can lead to better performance. Our PQR-based deep BIQA model also achieves standout results on the other databases.

\subsection{Cross Database Evaluations}

In our final reported series of experiments, we conducted cross database evaluations to compare the generalizabilities of the learned PQR and SQR based models using the three studied CNN architectures. In each case, one of the four databases was used for training, then the learned models were tested on the other three databases. The SRCC results are reported in Table \ref{table:cross database}.

From Table \ref{table:cross database} it may be observed that, although PQR delivers improved cross database evaluation performance relative to SQR in most cases, the achieved cross database prediction performance is usually poor. Specifically, the models trained on the LIVE Challenge database achieve unsatisfactory results on all the synthetic databases, and vice versa. The best cross database results are obtained between the LIVE IQA and CSIQ databases, whose distortion compositions are very similar. The poor cross database evaluation performance between LIVE Challenge and the three synthetic distortion databases is a common problem for all of the
studied deep BIQA methods.

\section{Conclusion and Discussion}

We were able to train deep BIQA models using our probabilistic quality representation (PQR) to accurately predict image quality,
while achieving faster convergence with a greater degree of stability. By imposing probabilistic constraints on the learned prediction
mapping, we essentially regularize the learning process. Our extensive experiments on existing IQA databases demonstrated that deep models
trained using PQR were able to achieve the uniformly best quality prediction accuracy, especially on the LIVE Challenge database, which is composed of real-world images degraded by complex, multiple authentic distortions.
In our view, this is largely a by-product of the fact that AlexNet and ResNet50 were trained on real-world images containing authentic
 distortions.
It was found that deep models trained on synthetic databases did not perform very well on LIVE Challenge; and vice versa.
This suggests that image quality prediction models trained on synthetic databases may not necessarily be expected to
perform well in real-world practice. Since this is the bona fide goal of IQA algorithm design (rather than performing well on a given
database), we begin to wonder about the ultimate value of using simulated distortions, at least in isolation, as training vehicles.

%

{\small
\bibliographystyle{IEEEtran}
\bibliography{refs}
}

\end{document}